\documentclass{article}

\usepackage{microtype}
\usepackage{graphicx}
\usepackage{booktabs} %
\usepackage{enumitem}
\usepackage{graphicx}
\usepackage{caption}
\usepackage{subcaption}

\def\loose{\looseness=-1}

\usepackage[accepted]{style/icml2024}

\usepackage{amssymb}%
\usepackage{pifont}
\usepackage{amsmath}
\usepackage{amssymb}
\usepackage{mathtools}
\usepackage{amsthm}
\usepackage{xspace}
\usepackage[colorlinks=true, allcolors=blue]{hyperref}
\usepackage[capitalize,noabbrev]{cleveref}

\newcommand{\std}[1]{\textcolor{black}{\scriptsize{$\pm #1$}}}

\usepackage{amsthm}
\usepackage{xcolor}
\usepackage{amsmath,amsfonts,bm}
\makeatletter
\newtheorem*{rep@theorem}{\rep@title}
\newcommand{\newreptheorem}[2]{%
\newenvironment{rep#1}[1]{%
 \def\rep@title{#2 \ref{##1}}%
 \begin{rep@theorem}}%
 {\end{rep@theorem}}}
\makeatother

\newreptheorem{theorem}{Theorem}

\definecolor{myred}{RGB}{215,48,39}
\definecolor{mygreen}{RGB}{26,152,80}

\newcommand{\halfmark}{\textcolor{gray}{\checkmark\kern-1.1ex\raisebox{.7ex}{\rotatebox[origin=c]{125}{--}}}}
\usepackage[framemethod=TikZ]{mdframed}
\mdfdefinestyle{MyFrame}{%
    linecolor=black,
    outerlinewidth=.3pt,
    roundcorner=5pt,
    innertopmargin=1pt, %
    innerbottommargin=1pt, %
    innerrightmargin=1pt,
    innerleftmargin=1pt,
    backgroundcolor=black!0!white}

\mdfdefinestyle{MyFrame2}{%
    linecolor=white,
    outerlinewidth=1pt,
    roundcorner=1pt,
    innertopmargin=0,
    innerbottommargin=0,
    innerrightmargin=7pt,
    innerleftmargin=7pt,
    backgroundcolor=black!3!white}

\mdfdefinestyle{MyFrameEq}{%
    linecolor=white,
    outerlinewidth=0pt,
    roundcorner=0pt,
    innertopmargin=0pt,
    innerbottommargin=0pt,
    innerrightmargin=7pt,
    innerleftmargin=7pt,
    backgroundcolor=black!3!white}

\newcommand{\RNum}[1]{\uppercase\expandafter{\romannumeral #1\relax}}

\newcommand{\E}{\mathcal{E}}

\newcommand{\R}{\mathcal{R}}

\newcommand{\vertiii}[1]{{\left\vert\kern-0.25ex\left\vert\kern-0.25ex\left\vert #1 
    \right\vert\kern-0.25ex\right\vert\kern-0.25ex\right\vert}}
\newcommand{\vertiiii}[1]{{\vert\kern-0.25ex\vert\kern-0.25ex\vert #1 
    \vert\kern-0.25ex\vert\kern-0.25ex\vert}}

\usepackage{mathtools}

\newcommand{\xhdr}[1]{{\noindent\bfseries #1}.}
\newcommand{\cut}[1]{}

\newcommand{\removelatexerror}{\let\@latex@error\@gobble}

\def\eqref#1{Eq.~\ref{#1}}

\def\1{\bm{1}}

\def\eps{{\epsilon}}

\DeclareMathAlphabet{\mathsfit}{\encodingdefault}{\sfdefault}{m}{sl}
\SetMathAlphabet{\mathsfit}{bold}{\encodingdefault}{\sfdefault}{bx}{n}

\def\gB{{\mathcal{B}}}

\def\gE{{\mathcal{E}}}

\def\gL{{\mathcal{L}}}

\def\gN{{\mathcal{N}}}

\def\gS{{\mathcal{S}}}

\def\gW{{\mathcal{W}}}

\def\gZ{{\mathcal{Z}}}

\def\sE{{\mathbb{E}}}

\def\R{{\mathbb{R}}}

\newcommand{\sethree}{\mathrm{SE(3)}}

\newcommand{\name}{i\textsc{DEM}\xspace}
\newcommand{\dem}{\textsc{DEM}\xspace}
\newcommand{\pdem}{p\textsc{DEM}\xspace}
\newcommand{\namelong}{\textsc{Iterated Denoising Energy Matching}\xspace}

\makeatletter
\renewcommand*{\appendixautorefname}{\S\@gobble}
\renewcommand*{\sectionautorefname}{\S\@gobble}
\renewcommand*{\subsectionautorefname}{\S\@gobble}
\renewcommand*{\subsubsectionautorefname}{\S\@gobble}
\makeatother

\newcommand{\ie}{\textit{i.e.}}
\newcommand{\eg}{\textit{e.g.}}

\usepackage[createShortEnv,conf={no link to proof, text link},commandRef=autoref]{proof-at-the-end}

\usepackage[textsize=tiny]{todonotes}

\begin{document}

\twocolumn[
\icmltitle{Iterated Denoising Energy Matching for Sampling from Boltzmann Densities}

\icmlsetsymbol{equal}{*}

\begin{icmlauthorlist} 
\icmlauthor{Tara Akhound-Sadegh}{equal,mila,mcgill,df}
\icmlauthor{Jarrid Rector-Brooks}{equal,mila,df,udem}
\icmlauthor{Avishek Joey Bose}{equal,mila,df,oxford}
\icmlauthor{Sarthak Mittal}{mila,udem}
\icmlauthor{Pablo Lemos}{mila,df,udem,ciela,cca}
\icmlauthor{Cheng-Hao Liu}{mila,mcgill,df}
\icmlauthor{Marcin Sendera}{mila,udem,ju}
\icmlauthor{Siamak Ravanbakhsh}{mila,mcgill,cifar}
\icmlauthor{Gauthier Gidel}{mila,udem,cifar}
\icmlauthor{Yoshua Bengio}{mila,udem,cifar}
\icmlauthor{Nikolay Malkin}{mila,udem}
\icmlauthor{Alexander Tong}{mila,df,udem}
\end{icmlauthorlist}

\icmlaffiliation{mila}{Mila -- Qu\'ebec AI Institute}
\icmlaffiliation{udem}{Universit\'e de Montr\'eal}
\icmlaffiliation{oxford}{University of Oxford}
\icmlaffiliation{mcgill}{McGill University}
\icmlaffiliation{ciela}{Ciela Institute}
\icmlaffiliation{ju}{Jagiellonian University}
\icmlaffiliation{cca}{Center for Computational Astrophysics, Flatiron Institute}
\icmlaffiliation{cifar}{CIFAR}
\icmlaffiliation{df}{Dreamfold}

\icmlcorrespondingauthor{}{\{jarrid.rector-brooks,tara.akhoundsadegh\}@mila.quebec}

\icmlkeywords{Machine Learning, ICML}

\vskip 0.3in
]

\printAffiliationsAndNotice{\icmlEqualContribution} %

\begin{abstract}
\loose
Efficiently generating statistically independent samples from an unnormalized probability distribution, such as equilibrium samples of many-body systems, is a foundational problem in science.
In this paper, we propose \namelong (\name), an iterative algorithm that uses a novel stochastic score matching objective leveraging solely the energy function and its gradient---and no data samples---to train a diffusion-based sampler. Specifically, \name alternates between (I) sampling regions of high model density from a diffusion-based sampler and (II) using these samples in our stochastic matching objective to further improve the sampler. 
\name is scalable to high dimensions as the inner matching objective, is \emph{simulation-free}, and requires no MCMC samples.
Moreover, by leveraging the fast mode mixing behavior of diffusion, \name smooths out the energy landscape enabling efficient exploration and learning of an amortized sampler. 
 We evaluate \name on a suite of tasks ranging from standard synthetic energy functions to
 invariant $n$-body particle systems. We show that the proposed approach achieves state-of-the-art performance on all metrics and trains $2-5\times$ faster, which allows it to be the first method to train using energy on the challenging $55$-particle Lennard-Jones system.

\end{abstract}

\section{Introduction}
\label{sec:introduction}
\begin{figure}[t]
    \includegraphics[width=\linewidth]{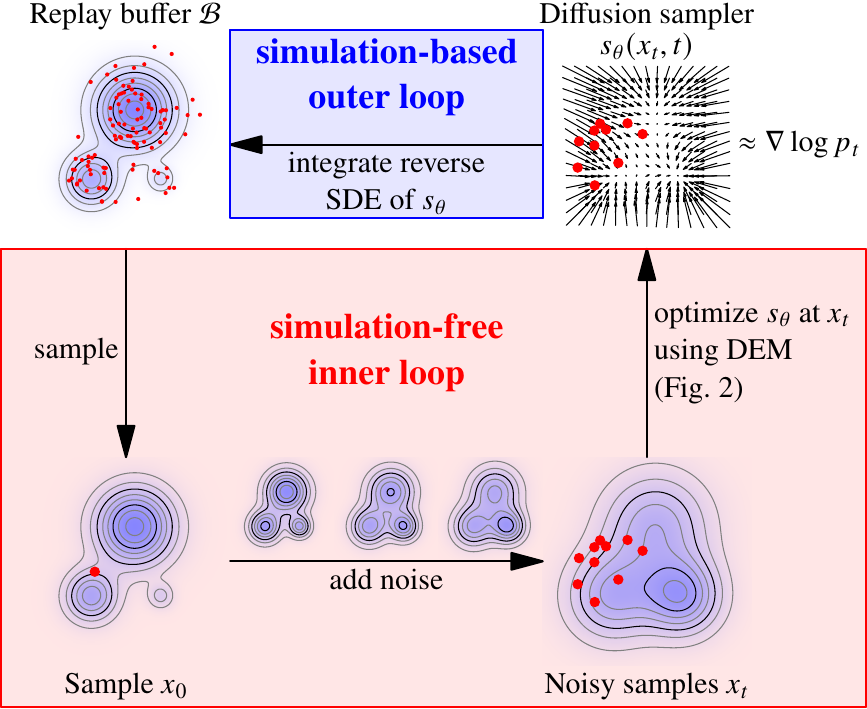}
    \caption{
        \loose \name fits a diffusion sampler to a target distribution given by an unnormalized density. In the outer loop, \name populates a buffer with samples from the current model $s_\theta$. In the inner loop, \name uses the \dem objective (\autoref{sec:denoising_boltzmann_target}) to regress $s_\theta$ to an estimate of the score at noised samples from the buffer. The inner loop is simulation-free, \ie, requires no SDE integration.
    }
    \label{fig:visual_abstract}
\end{figure}

\loose
A fundamental task in probabilistic inference is drawing samples from an unnormalized probability density.
Computational approaches have been employed to tackle this significant problem, yielding a multitude of applications across various scientific domains,
such as spin-lattice states~\citep{li2018neural,nicoli2020asymptotically}, nuclear physics~\citep{albergo2019flow}, and proteins~\citep{jumper2021highly,bose2023se}. In this work, we focus on sampling from the (target) equilibrium distribution  $\mu_{\text{target}}$ of many-body systems, \eg, molecules,  with density proportional to a Boltzmann-type distribution $\mu_{\text{target}}(x) \propto \exp(-\mathcal{E}(x))$ with the specific goal of efficiently drawing samples that cover all modes of the complex and dimensionless energy $\mathcal{E}$. Moreover, the density $\mu_{\text{target}}$ associated with the system is symmetric as the energy $\gE$ function is invariant to rotations, reflections, and permutations of particles in $3\text{D}$ space. 

\loose
Unlike the typical machine learning settings where the starting point is a dataset, learning to sample in many of these scientific settings is especially challenging as we often have little to no initial samples from $\mu_{\text{target}}$, or the given samples only cover a small set of modes~\citep{noe2019boltzmann}. Acquiring additional high-quality samples can be achieved by leveraging Monte Carlo (MC) techniques such as Annealed Importance Sampling~\citep[AIS;][]{neal2001annealed} or Sequential Monte Carlo~\citep[SMC;][]{del2006sequential} or simulating the actual (molecular) dynamics~\citep[MD;][]{leimkuhler2013rational}. Unfortunately, MC techniques and MD are computationally expensive with poor scaling to high dimensions inhibiting their easy application in complex high-dimensional physical systems. 

\loose
The lack of sufficient data in sampling from Boltzmann-type distributions also precludes training deep generative models, $q_{\theta}$, to match $\mu_{\text{target}}$ in the classical sense by maximizing the likelihood---\ie, minimizing $\text{KL}(\mu_{\text{target}} \lvert \rvert  q_{\theta})$. An alternative to Markov Chain Monte Carlo (MCMC), SMC, and MD is to consider variational approaches where $q_{\theta}$ is optimized using a metric where the samples are drawn under the \emph{model} rather than the data. A convenient choice is the reverse-KL divergence, $\text{KL}(q_{\theta} \lvert \rvert \mu_{\text{target}})$, but such a discrepancy measure suffers from mode-seeking behavior and is incapable of exploring the entire energy landscape. In scientific applications, one avenue is to leverage exact likelihood-based generative models---\eg, normalizing flows~\citep{rezende2015variational,dinh2016density}---trained using a combination of the forward and reverse KL to approximate $\mu_{\text{target}}$ and use importance sampling weights to correct for modelling errors~\citep{noe2019boltzmann,midgley2022flow}. Despite the popularity of this approach, termed Boltzmann generators, the efficacy of the sampled points under the target is underpinned by the quality of initial samples from MCMC or MD, the expressive power of the class of flows, as well as the fidelity of the importance sampling estimator. 

\loose
Given the complexity of physical processes, purely learning-based neural samplers such as the Path Integral Sampler~\citep[PIS;][]{zhang2021path}, Time-reversed Diffusion Sampler~\citep[DIS;][]{berner2022optimal}, and Denoising Diffusion Sampler~\citep[DDS;][]{vargas2023denoising} are attractive substitutes for Boltzmann generators as they 
can amortize MCMC, and learn in the absence of data. Despite this, at present all neural samplers must \emph{simulate} expensive forward and reverse trajectories and their gradients during learning which prevents their use when scaling to large-scale scientific applications. Thus, the search for an improved neural sampler motivates the following research question:

\hangindent=0.3cm
\hspace*{0.1cm} \
\textit{Can we find a scalable sampler that can learn from $\mathcal{E}(x)$ and $\nabla \gE$ while achieving high mode-coverage of $\mu_{\text{target}}$?}

\xhdr{Present work}
\loose
In this paper, we propose \namelong (\name) a neural sampler based on denoising diffusion models for sampling from a Boltzmann distribution with a known energy function. \name is not only computationally tractable (\autoref{tab:summary}), but also provides a good coverage of all modes of the distribution. In addition, \name can readily be imbued with any symmetries that manifest as invariances in $\gE(x)$ making it well suited for scientific applications.
Furthermore, in stark contrast to methods using MCMC, variational objectives, AIS, FAB, and SMC~\citep{noe2019boltzmann,midgley2022flow,matthews2022continual} \name uses diffusion sampled data from the model mixed with an exploratory off-policy scheme to avoid the need for samples from $\mu_{\text{target}}$ while providing the option of using existing data.

\loose
Our proposed approach \name is structured as a bi-level algorithm in which the inner loop iteratively updates a diffusion sampler using a novel \emph{simulation-free} stochastic regression objective directly on the energy function $\E(x)$. The outer loop of \name uses simulation of the reverse SDE of the updated (iterated) diffusion sampler and serves two important goals: 1.) it amortizes sampling---imitating a well-mixed MCMC chain as training progresses and 2.) it enables efficient exploration of the energy landscape as the inner loop updates push the model closer to matching $\mathcal{E}$, allowing us to sample closer to the true energy.
Intuitively, as depicted in \autoref{fig:visual_abstract}, \name takes inspiration from denoising objectives popularized in conventional diffusion models~\citep{ho2020denoising} and constructs a forward Gaussian process that adds noise in the (energy) function space until we reach the unnormalized log probability of a standard Normal distribution. By smoothing the energy directly using diffusion, \name builds upon important theoretical benefits such as fast-mixing times in high dimensions~\citep{de2021diffusion}. Reaching all modes during inner loop updates provides an informative learning target for the iterated diffusion sampler, whose reverse process learns to then transport particles from low to high-density regions under $\mu_{\text{target}}$. %

\loose
We test the empirical caliber of \name by conducting a range of experiments on synthetic Gaussian mixtures as well as $\sethree \times \mathbb{S}_n$-invariant double-well and Lennard-Jones potentials associated to $n$-body particle systems with DW-4, LJ-13, and LJ-55~\citep{kohler2020equivariant,klein2023equivariant}. We empirically find that \name achieves performance which is competitive and often exceeds previous state-of-the-art approaches in FAB~\citep{midgley2022flow} and all neural sampler baselines~\citep{zhang2021path,vargas2023denoising}. Importantly, the performance of \name is achieved at a fraction of the training and memory cost of previous approaches which enables {\bf \name to be the \emph{first} method to successfully scale to LJ-55} using energy-based training.

\definecolor{myred}{RGB}{215,48,39}
\definecolor{mygreen}{RGB}{26,152,80}
\newcommand{\yes}{\textcolor{mygreen}{\ding{51}}}
\newcommand{\no}{\textcolor{myred}{\ding{55}}}
\newcommand{\maybe}{\textcolor{gray}{\checkmark\kern-1.1ex\raisebox{.7ex}{\rotatebox[origin=c]{125}{--}}}}
\begin{table}[t]
    \centering
    \caption{A comparison of approaches that are a) MCMC-free, b) are trained off-policy, c) require $L$ forward simulation steps while training, and d) require backward gradients through time for $d$-dimensional samples. See \S\ref{app:related_work} for details and discussion.}
    \label{tab:summary}
    \resizebox{\linewidth}{!}{%
    \begin{tabular}{@{}lcccc}
    \toprule
    Method & MCMC-free & Off-policy & Time & Memory\\
    \midrule
    FAB \citep{midgley2022flow}  &\no  & \yes & $\mathcal{O}(L)$ & $\mathcal{O}(L+d)$\\
    PIS \citep{zhang2021path} &\yes & \no & $\mathcal{O}(L)$ & $\mathcal{O}(L d)$\\
    DDS \citep{vargas2023denoising} &\yes & \no & $\mathcal{O}(L)$ & $\mathcal{O}(L d)$\\
    \pdem (ours) & \yes & \yes & $\mathcal{O}(1)$ & $\mathcal{O}(d)$ \\
    \name (ours) &\yes & \maybe  & $\mathcal{O}(L)$ & $\mathcal{O}(d)$\\

    \bottomrule
    \end{tabular}
    }
\end{table}

\section{Background and preliminaries}
\label{sec:background}

We are concerned with sampling problems in which we seek to draw samples from a target distribution $\mu_{\text{target}}$ over $\R^d$, %
\begin{equation*}
    \mu_{\text{target}}(x) = \frac{\exp\left( -\E(x) \right)}{\mathcal{Z}}, \ \mathcal{Z} = \int_{\mathbb{R}^d} \exp\left( -\E(x) \right) dx.
\end{equation*}
\loose
The denominator $\mathcal{Z}$ is known as the \emph{partition function} and is intractable for general energies $\E$. Consequently, we are unable to evaluate the exact density at a point $x \in \mathbb{R}^d$. Instead, we assume that we have access to the energy $\E:\R^d\to\R$ and thus to the unnormalized probability density, $\mu_{\text{target}}\propto \exp(-\E(x))$. In scientific applications, such densities---modeled as \emph{Boltzmann distributions}---can be used to express the probability of a system being in a particular state as a function of an energy function $\E(x)$. We next outline various standard approaches to sampling from $\mu_{\text{target}}$.

\subsection{Classical sampling methods}
\label{sec:mcmc}

\loose
It is often the case we wish to compute expectations of some observable $f(x)$ by drawing samples from our distribution of interest $x \sim \mu_{\text{target}}$. If $\mu_{\text{target}}$ is an easy-to-sample distribution we could simply compute the Monte Carlo estimate which is the sample average. But, if $\mu_{\text{target}}$ is complex or not easy to sample from we must resort to alternative methods. 

\xhdr{Importance sampling}
\loose
By selecting an easy-to-sample from distribution $q(x)$ it is possible to construct a consistent estimator. We do so by drawing $K$ independent samples $x^i \sim q(x), i \in [K]$ and computing the importance weights which is the ratio $w(x^i) = \exp(-\E(x^i)) / q(x^i)$. This allows us to estimate the expectation of $f(x)$ under $\mu_{\text{target}}$ as:
\begin{equation*}
    \text{IS}:= \mathbb{E}_{x \sim \mu_{\text{target}}}[f(x)] \approx \frac{\sum_k w(x^i)f(x^i)}{\sum_k w(x^i)}, \ x^i \sim q(x).
 \end{equation*}
 \loose
The optimal $q(x^i)$ is the one that minimizes the variance of the estimator and is roughly proportional to $f(x^i)\mu_{\text{target}}(x^i)$~\citep{mcbook}. As a result, finding a good $q$ in high dimensions or when $\mu_{\text{target}}$ is multimodal with separated modes is often challenging.

\loose
A detailed review of MCMC techniques is provided in \autoref{app:sampling_app}.

\subsection{Denoising diffusion}
\label{sec:denoising_diffusion}

\loose
Diffusion models~\citep{sohl2015deep,ho2020denoising,song2020score} are probabilistic models whose generative process is the reverse of a tractably sampled stochastic process. Our \name borrows key modeling assumptions from diffusion, and we briefly review them here.

\loose
We denote $p_t$ with $t\in [0,1]$ the marginal distribution of the diffusion process which starts at $p_0=\mu_{\text{target}}$ as a distribution over $\R^d$. In typical denoising diffusion, $p_0$ is a mixture of Dirac measures over the training dataset. We consider the stochastic differential equation (SDE),
\begin{equation}
    dx_t = -\alpha(t) x_t\,dt + g(t)\,dw_t,
    \label{eq:ou}
\end{equation}
where $w_t$ is a Brownian motion and $\alpha$ and $g$ are functions of time. This SDE is known as the \emph{forward (noising) process} which progressively adds noise starting from data $x_0 \sim p_0$ and runs over an interval $t\in[0,1]$. Common choices for the decay rate $\alpha$ include $\alpha(t)=0$ (variance-exploding (VE)) and $\alpha(t)=\frac{g(t)^2}{2}$ (variance-preserving (VP)). 

\loose
The marginal distribution of the process (\ref{eq:ou}) at time $t$ is denoted $p_t$ and has a smooth density for $t>0$ under mild assumptions on $\mu_{\text{target}}$. The corresponding \emph{reverse process} SDE with Brownian motion $\overline{w}_t$ associated with (\ref{eq:ou}) is then
\begin{equation}
    dx_t = [-\alpha(t) x_t - g(t)^2\nabla\log p_t(x_t)]\,dt + g(t)\,d\overline{w}_t.
    \label{eq:reverse_sde}
\end{equation}
\loose
To use the reverse SDE as a generative model, it is necessary to estimate the (Stein) score function of the convolved data distribution, $\nabla\log p_t(x_t)$. Denoising diffusion models fit a neural network $s_\theta(x_t,t)$, to this score via a stochastic regression. To be precise, in the example of the VE SDE, the density $p_t$ is recognized as a convolution:
\begin{equation}
    p_t=p_0*\gN(0,\sigma^2_t),\quad \sigma^2_t:=\int_0^tg(s)^2\,ds,
    \label{eq:pt_as_convolution}
\end{equation}
\hbox{from which one can derive that}\\[-1em]
\begin{equation}
    \nabla_{x_t}\log p_t(x_t) = \sE_{x_0\sim p(x_0\mid x_t)}\overbracket{\left[\frac{x_0-x_t}{\sigma^2_t}\right]}^{\hspace*{-5mm}=\nabla_{x_t}\log\gN(x_t;x_0,\sigma^2_t)\hspace*{-5mm}},
    \label{eq:score_as_expectation}
\end{equation}
where $p(x_0|x_t)\propto p(x_0)p(x_t|x_0)=p(x_0)\gN(x_t;x_0,\sigma^2_t)$. This expression suggests a stochastic regression objective---called denoising score matching---for the estimated score:
\begin{equation}
    \gL=\sE_{\substack{x_0\sim p_0(x_0)\\x_t\sim\gN(x_t;x_0,\sigma^2_t)}}\left\|\frac{x_0-x_t}{\sigma_t^2}-s_\theta(x_t,t)\right\|^2.
    \label{eq:score_regression}
\end{equation}
The objective (\ref{eq:score_regression}) requires sampling from $p_0$ to be tractable. In the next section, we will study the case where $p_0$ is not tractable but is a Boltzmann density with known energy $\gE$.

\section{\namelong}
\label{sec:method}

\loose
We now present \name, which is designed to sample from a distribution $\mu_{\text{target}}$. From henceforth, we will interchangeably use $\mu_{\text{target}}$ and $p_0$ to refer to the target density at time $t=0$ and set $p_1$ to denote a tractable prior at time $t=1$.  We assume that $\gE$ is known and $\nabla \gE$ is cheaply computable, but that $\gZ$ is not, and thus exact sampling is not tractable. 

\loose
We motivate the design of \name by first outlining the two principal challenges that inhibit the training of a diffusion sampler in the absence of data.  \textbf{(C1)} The score function $\nabla \log p_t(x_t)$ is not available and \textbf{(C2)} we do not know where in the sample space to match the score.

\loose
To overcome these challenges \name is composed of two key algorithmic components organized in a bi-level iterative scheme. The inner loop tackles $\mathbf{(C1)}$ by proposing \textsc{Denoising Energy Matching} (\dem)  a novel stochastic regression objective to the score using only the energy $\gE$ and its gradient while the outer loop addresses $\mathbf{(C2)}$ by proposing informative starting points $x_0$ which can then be diffused and used in the subsequent inner loop of the algorithm.

\loose
\begin{enumerate}[label=\textbf{(C\arabic*)},left=0pt,nosep]
\item 
\xhdr{Inner Loop} The sampler $s_{\theta}$ is trained to approximate the score of the target density convolved with varying levels of noise. Specifically, $s_{\theta}$ is updated using \dem  (\autoref{sec:denoising_boltzmann_target}). In principle, we can optimize $s_\theta$ with respect to the \dem objective at any point in time $t$ and point $x_t$, but an optimal training scheme would prudently select points $x_t$ at which to train the score estimator. 
The `off-policy' nature of the \dem objective allows flexibility in the choice of $x_t$.
\item
\xhdr{Outer Loop} For the \dem objective to provide a useful learning signal it is critical to select informative points $x_t$. While any sampling strategy is possible, including off-policy methods and MCMC, we make an algorithmic choice to utilize $s_{\theta}$ via its reverse SDE as an amortized sampler (\autoref{sec:outer_loop_sampling}), whose proposals enable fast exploration in high dimension. Iteratively updating $s_{\theta}$ in every inner loop phase synergistically improves the sampling quality of $s_{\theta}$ in the outer loop.
\end{enumerate}

\loose
A complete description of \name is provided in \cref{alg:idem}.

\subsection{Denoising diffusion with a Boltzmann target (C1)}
\label{sec:denoising_boltzmann_target}

We consider the same noising process as in \autoref{sec:denoising_diffusion}, given by an SDE of the form (\ref{eq:ou}), but now with $p_0$ being a Boltzmann density. Recall from (\ref{eq:reverse_sde}) that reversing the noising process requires the score function $\nabla\log p_t(x_t)$, where $p_t=p_0*\gN(0,\sigma^2_t)$. However, unlike in the case of an empirical data distribution $p_0$, we cannot tractably sample $p_t$ or regress to its score. The main ideas of the stochastic regression objective in \name are (1) to estimate the score of $p_t$ by Monte Carlo and (2) to regress a neural network estimator $s_{\theta}$ to this estimated score. We describe each idea in turn. %

\loose
\xhdr{MC score estimation} We write the score of $p_t$ as an expectation in a manner similar to (\ref{eq:score_as_expectation}), here for the VE SDE:
\begin{equation*}
    \nabla\log p_t(x_t) 
    = \frac{\nabla\left(p_0*\gN(0,\sigma^2_t)\right)(x_t)}{p_t(x_t)}.
\end{equation*}
The key observation is that the gradient of the Gaussian convolution with $p_0$ can be done in a specific way that gives an avenue for efficient estimation as described below.
\begin{align}
    \nabla\log p_t(x_t) 
    &= \frac{((\nabla p_0)*\gN(0,\sigma^2_t))(x_t)}{p_t(x_t)} \label{eq:score_as_convolution} \\
    &= \frac{\sE_{x_{0\mid t}\sim\gN(x_t,\sigma^2_t)}[\nabla p_0(x_{0\mid t})]}{\sE_{x_{0\mid t}\sim\gN(x_t,\sigma^2_t)}[p_0(x_{0\mid t})]} \nonumber \\
    &= \frac{\sE_{x_{0\mid t}\sim\gN(x_t,\sigma^2_t)}[\nabla \exp(-\gE(x_{0\mid t}))]}{\sE_{x_{0\mid t}\sim\gN(x_t,\sigma^2_t)}[\exp(-\gE(x_{0\mid t}))]},
    \label{eq:score_as_expectation_boltzmann}
\end{align}
where (\ref{eq:score_as_convolution}) is by a standard property of convolutions and (\ref{eq:score_as_expectation_boltzmann}) uses that the normalizing factor $\frac1\gZ$ appears in both the numerator and denominator.
Since, (\ref{eq:score_as_expectation_boltzmann})
is true for \emph{any} $x_t$, it means that it can provide a training signal to learn the score function using samples that come from any distribution, not necessarily those associated with $\mu_{\rm target}$. This provides two principal advantages: simulation-free computation of the gradient, and off-policy training, which can be exploratory.

\loose
We note a connection between (\ref{eq:score_as_convolution}) and the score in the empirical case (\ref{eq:score_as_expectation}). In (\ref{eq:score_as_expectation}), the gradient is placed on the second term of the convolution, $\gN(0,\sigma^2_t)$, which allows estimation when $p_0$ has no density but is tractable to sample. In (\ref{eq:score_as_convolution}), the gradient is instead placed on the first term, $p_0$, taking advantage of the fact that sampling from the normal distribution is feasible, while for $p_0$ sampling is not possible but we can compute a gradient.~\autoref{app:fm} contains further discussion and connection with flow matching algorithms, stochastic control, and the recently proposed Reverse Diffusion Monte Carlo \citep{huang2023reverse}.

The expression (\ref{eq:score_as_expectation_boltzmann}) suggests a Monte Carlo estimator that uses the same set of samples from $\gN(x_t,\sigma^2_t)$ to approximate the numerator and denominator. That is, we write
\begin{align}
    \nabla\log p_t(x_t)
    &\approx \frac{\frac1K\sum_i\nabla\exp(-\gE(x_{0\mid t}^{(i)}))}{\frac1K\sum_i\exp(-\gE(x_{0\mid t}^{(i)}))}\nonumber\\
    &=\nabla_{x_t}\log\sum_i\exp(-\gE(x_{0\mid t}^{(i)})), \label{eq:mc_score_esimate}\\
    x_{0\mid t}^{(1)},\dots,x_{0 \mid t}^{(K)} & \sim\gN(x_t,\sigma^2_t),\nonumber
\end{align}
\loose
where in the second line $x_{0 \mid t}^{(i)}$ is understood as a function of $x_t$ via the reparametrization $x_{0 \mid t}^{(i)}=x_t+\epsilon^{(i)}$, $\epsilon^{(i)}\sim\gN(0,\sigma^2_t)$ and the notation $0 \mid t$ indicates a sample at time $t=0$ is drawn from a distribution centred at $x_t$. For numerical stability in low-density regions (\ref{eq:mc_score_esimate}) is implemented using the $\rm LogSumExp$ trick.

\begin{figure}[t]
    \centering
    \includegraphics[width=1.0\linewidth]{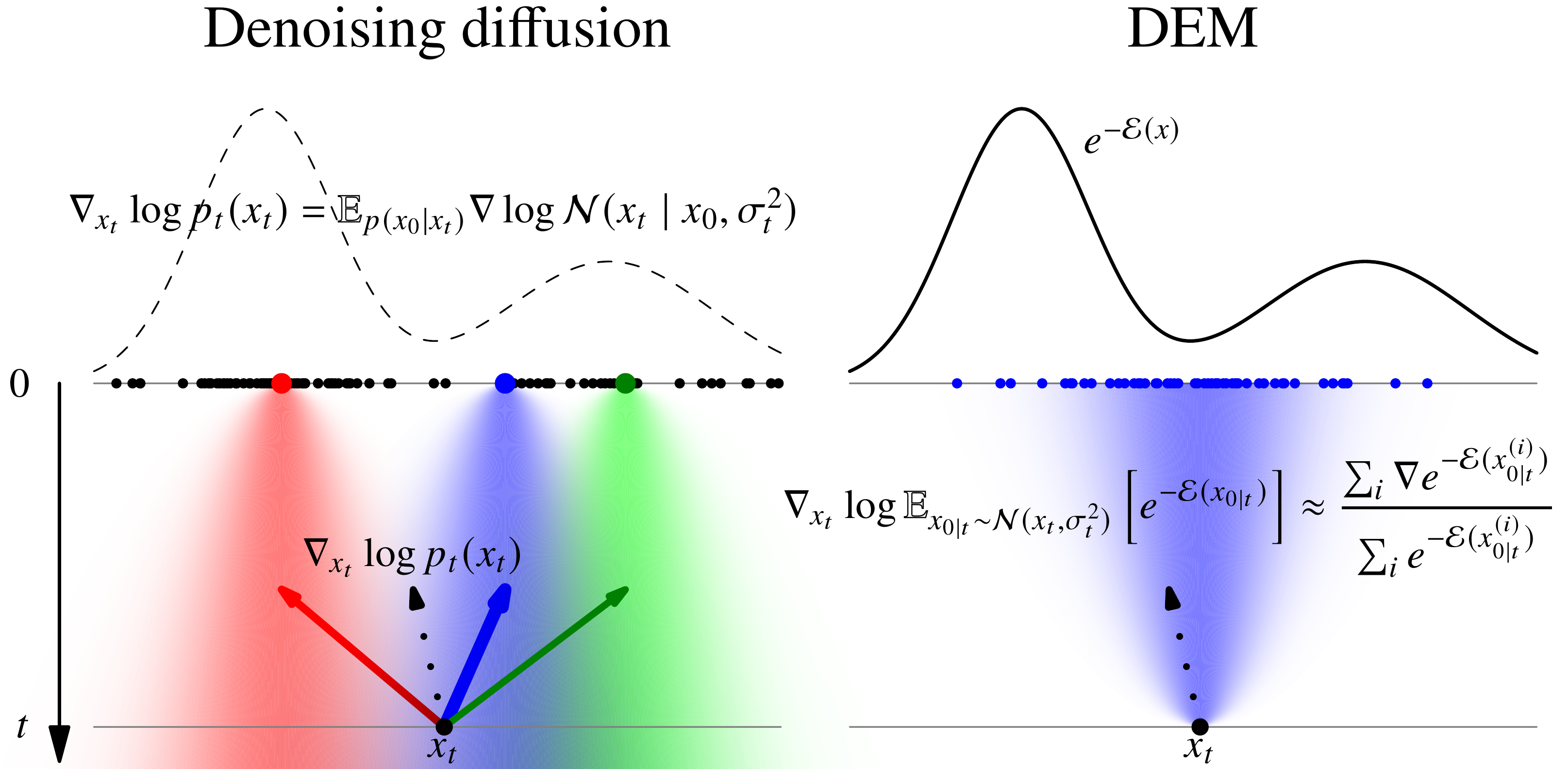}
    \caption{
        \loose Two ways of estimating the score $\nabla\log p_t(x_t)$. \textbf{Left:} A diffusion model estimates the score convolved with noise by stochastically regressing to the scores of distributions conditioned on $x_0$---\ie, points \textcolor{red}{$\bullet$}, \textcolor{green}{$\bullet$}, \textcolor{blue}{$\bullet$}---weighted by the likelihood of $p(x_0 | x_t)$ (indicated by the arrow thickness). This regression requires samples from $\mu_{\text{target}}$. \textbf{Right:} \dem assumes an unnormalized density over $x_0$ and expresses the score of the convolved density as an expectation and regresses to a consistent estimator of this score.
    }
    \label{fig:ddpm_vs_idem}
\end{figure}

The $K$-sample approximation in (\ref{eq:mc_score_esimate}), which we denote $\gS_K(x_t,t)$, can also be understood as an importance-weighted estimate over $p_0(x_{0\mid t})\gN(x_{0\mid t};x_t,\sigma^2_t)$ as follows, 
\begin{align}
    \gS_K(x_t,t)&=-\sum_iw_i\nabla\gE(x_{0\mid t}^{(i)}),\label{eq:iw_score_estimate}\\
    w_i&:=\frac{\exp(-\gE(x_{0\mid t}^{(i)}))}{\sum_j\exp(-\gE(x_{0\mid t}^{(j)}))}\underset{i}{\propto} p_0(x_{0\mid t}^{(i)}).\nonumber
\end{align}
\loose
This recalls the expectation over $p(x_0\mid x_t)$ in (\ref{eq:score_as_expectation}). In addition,
\autoref{fig:ddpm_vs_idem} visually illustrates the MC estimator in (\ref{eq:mc_score_esimate}), which is distinguished from a classical diffusion objective.

\loose
The estimator $\gS_K(x_t,t)$ is a consistent estimator and we characterize its bias with the following proposition.

\begin{propositionE}[][end,restate]
If $\exp(-\gE(x_{0\mid t}^{(i)}))$  and $\|\nabla\exp (-\gE(x_{0\mid t}^{(i)}))\|$ are sub-Gaussian, then, there exists a constant $c(x_t)$ such that with probability $1-\delta$ (over $x_{0\mid t}^{(i)}\sim \gN(x_t,\sigma_t^2)$) we have $\|\gS_K(x_t,t) - \nabla \log p_t (x_t) \| \leq \frac{c(x_t)\log\left(\tfrac{1}{\delta}\right)}{\sqrt{K}}\,.$

\label{prop:bias}
\end{propositionE}
\begin{proofE}
    We seek to estimate $\nabla_{x_t} \log p_t(x_t) = -\nabla \gE_t(x_t) $ where $\exp(-\mathcal{E}_t(x)) := \mathbb{E}_{x_{0\mid t} \sim \mathcal{N}(x_t, \sigma^2_t)} [ \exp(-\mathcal{E}(x_{0\mid t})) ] \propto  p_t(x) $.  With a slight abuse of notation let us denote $x_{0\mid t} = (x_{0\mid t}^{(1)}, \cdots, x_{0\mid t}^{(K)})$, we consider the \emph{biased estimator} $\gS_K(x_{0\mid t},t) = \nabla \log \frac{1}{K}\sum_i  \exp (-\gE(x_{0\mid t}^{(i)}))$ with $x_{0\mid t}^{(i)} \sim \mathcal{N}(x_t, \sigma^2_t)$. Denote, $\gS(x_t,t) := - \nabla \gE_t(x_t) $, we now estimate the bias of this estimator as a function of $K$.

If we assume that the random variables $\exp (-\gE(x_{0\mid t}^{(i)}))$ and $\nabla \exp (-\gE(x_{0\mid t}^{(i)}))$ with $x_{0\mid t}^{(i)}\sim \gN(x_t,\sigma_t^2)$ are sub-Gaussian, then by Hoeffding's inequality on sub-Gaussian random variables~\citep{vershynin2018high}, we have that there exists a constant $C>0$ such that  for any $\delta >0$ with probability $1-\delta$ we have
\begin{equation}
    \left|\frac1K\sum_i\exp(-\gE(x_{0\mid t}^{(i)}))-\exp(-\gE_t(x_t))\right| \leq C\sqrt{\frac{\log(\frac{2}{\delta})}{K}}
\end{equation}

and

\begin{equation}
     \left\|\frac1K\sum_i\nabla \exp(-\gE(x_{0\mid t}^{(i)}))-\nabla \exp(-\gE_t(x_t))\right\| \leq C\sqrt{\frac{\log(\frac{2}{\delta})}{K}}
\end{equation}

Thus, for $K \geq 4C^2 \log(1/\delta)\exp(2\gE_t(x_t))$ , with probability $1-\delta$ (over the sampling of $x_{0\mid t}$) we get,

\begin{align}
\|\gS_k(x_{0\mid t},t) - \gS(x,t)\|
   &=    \left \| \frac{\frac1K\sum_i\nabla\exp(-\gE(x_{0\mid t}^{(i)}))}{\frac1K\sum_i\exp(-\gE(x_{0\mid t}^{(i)}))} - \frac{\nabla\exp(-\gE_t(x_t))}{\exp(-\gE_t(x_t))}
   \right\| \\
   &= \left \| \frac{\exp(-\gE_t(x_t))\frac1K\sum_i\nabla\exp(-\gE(x_{0\mid t}^{(i)})) -\frac1K\sum_i\exp(-\gE(x_{0\mid t}^{(i)})\nabla\exp(-\gE_t(x_t))}{\exp(-\gE_t(x_t))\frac1K\sum_i\exp(-\gE(x_{0\mid t}^{(i)}))}
   \right\| \\
   &= \left \| \frac{e^{-\gE_t(x_t)}(\frac1K\sum_i\nabla e^{-\gE(x_{0\mid t}^{(i)})}-\nabla e^{-\gE_t(x_t)}) + (e^{-\gE_t(x_t)}-\frac1K\sum_ie^{-\gE(x_{0\mid t}^{(i)})})\nabla e^{-\gE_t(x_t)}}{\exp(-\gE_t(x_t))\frac1K\sum_i\exp(-\gE(x_{0\mid t}^{(i)}))}
   \right\| \\
   & \leq  C\sqrt{\frac{\log(\frac{2}{\delta})}{K}}
  \frac{\exp(-\gE_t(x_t))+\|\nabla\exp(-\gE_t(x_t))\|}{\exp(-\gE_t(x_t))\frac1K\sum_i\exp(-\gE(x_{0\mid t}^{(i)}))}\\
  & \leq  C\sqrt{\frac{\log(\frac{2}{\delta})}{K}}
  \frac{1+\|\nabla \gE_t(x)\|}{\exp(-\gE_t(x_t))-  C\sqrt{\frac{\log(\frac{2}{\delta})}{K}}}\\
  & \leq  \frac{2C\sqrt{\log(\frac{2}{\delta})}(1+\|\nabla \gE_t(x_t)\|)(\exp(\gE_t(x_t)))}{\sqrt{K}}
\end{align}
where for the last inequality we used the fact that $K$ is large enough to have $\exp(-\gE_t(x_t))-  C\sqrt{\frac{\log(\frac{2}{\delta})}{K}} \geq \exp(-\gE_t(x_t))/2$
   Thus, there exists a constant $c(x_t)>0$ such that we have with probability $1-\delta$
\begin{equation}
    \|\gS_k(x_{0\mid t},t) - \gS(x_t,t)\| \leq \frac{c(x_t) \sqrt{\log(1/\delta)}}{\sqrt{K}}.
\end{equation}
Thus we have shown the conclusion of the proposition holds for $K$ sufficiently large, which easily implies the general case (with a possibly larger value of $c(x_t)$).
\end{proofE}

\loose
We present all proofs in \autoref{sec:proofs}.
\autoref{prop:bias} elucidates that the bias of $\gS_K$ decays at a rate of $O(1/\sqrt{K})$. Note that this means that for regions with large values of $\mathbb{E}_{x_{0\mid t} \sim \gN(x_t, \sigma^2_t)}[\exp(-\gE(x_{0\mid t}))]$ we can obtain an accurate estimate for modest values of $K$. In contrast, for low-density regions we need $K$ large such that $\frac{c(x_t)}{\sqrt{K}} \leq \mathbb{E}_{x_{0\mid t} \sim \gN(x_t, \sigma^2_t)}[\exp(-\gE(x_{0\mid t}))]$, which motivates the search for an informative starting sample $x_0$ and is the focus of~\autoref{sec:outer_loop_sampling}. Finally, note that the sub-Gaussian assumption is relatively mild and all our energies and their gradient norms studied in this paper satisfy this property.

\loose
\xhdr{Regressing to the estimate} As in standard diffusion models, we aim to fit a neural network $s_\theta(x_t,t)$ with parameters $\theta$ to the score $\nabla\log p_t(x_t)$. This is achieved by minimizing the regression loss:
\begin{equation}
    \gL_{\dem}(x_t,t):=\|\gS_K(x_t,t)-s_\theta(x_t,t)\|^2,
    \label{eq:idem_loss}
\end{equation}
at a given point $x_t$ and time $t$. As the estimator $\gS_K$ is stochastic, the optimal solution for (\ref{eq:idem_loss}) in the space of all values for $s_\theta(x_t,t)$ is $s_\theta^*(x_t,t)=\sE[\gS_K(x_t,t)]$, which by \autoref{prop:bias}, approaches the true score $\nabla\log p_t(x_t)$ as $K\to\infty$.

\loose
The objective (\ref{eq:idem_loss}) can be computed for a fixed $x_t$, and its global minimum in function space does not depend on the choice of $t$ and $x_t$ at which it is optimized (as long as the training distribution has full support). This property is in contrast to (\ref{eq:score_regression}), in which $x_t$ must be sampled from a distribution conditioned on a data point $x_0$. The flexibility in the choice of $t$ and $x_t$ in $\gL_{\dem}$ allows ``off-policy" methods that recycle points generated by past iterations of the model.

We also note that to construct \dem we made use of two Gaussian convolutions, namely $\gN(0, \sigma^2_t)$ and $\gN(x_t, \sigma^2_t)$. The first convolution is used to create a probability $p_t$ from which we draw samples $x_t$. Sampling from $p_t$ enables us to smooth the energy landscape via diffusion. The second convolution is used to construct the MC estimate of the score $\nabla \log p_t$, which is the regression target for $s_{\theta}$.
\begin{algorithm}[t]
  \caption{\namelong}
  \label{alg:scfm}
\begin{algorithmic}

\STATE {\bfseries Input:} Network $s_{\theta}$, Batch size $b$, Noise schedule $\sigma^2_t$, Prior $p_1$, Num.\ integration steps $L$, Replay buffer $\gB$, Max Buffer Size $|\gB|$ , Num.\ MC samples $K$.
\WHILE{Outer-Loop} 

\STATE $\{x_1\}^b_{i=1} \sim p_1(x_1)$
\STATE $\{x_0 \}^b_{i=1} \gets$ \texttt{sde\_int}$(\{x_1\}^b_{i=1} , s_\theta, L)$ \COMMENT{Sample}
\STATE $\gB$ = $(\gB \cup \{x_0 \}^b_{i=1})$ \COMMENT{Update Buffer $\gB$}
\WHILE{Inner-Loop} 
\STATE $x_0 \gets \gB$.\texttt{sample() }\COMMENT{Uniform sampling from $\gB$}
\STATE $t \sim \mathcal{U}(0, 1)$, $x_t \sim \mathcal{N}(x_0, \sigma^2_t)$

\STATE $\gL_{\dem}(x_t,t)=\|\gS_K(x_t,t)-s_\theta(x_t,t)\|^2$
\STATE $\theta \gets \texttt{Update}(\theta, \nabla_\theta \mathcal{L}_{\dem}$)%
\ENDWHILE
\ENDWHILE
\OUTPUT $s_\theta$
\end{algorithmic}
\label{alg:idem}
\end{algorithm}
\subsection{Amortized sampling with a diffusion sampler (C2)}
\label{sec:outer_loop_sampling}

\loose
The \dem loss introduced in \autoref{sec:denoising_boltzmann_target} serves as a useful learning target whenever a sample $x_0$ corresponds to a high value of $\exp(-\gE(x_0))$. Specifically, constructing a stochastic regression objective starting from such an $x_0$ enables us to train $s_{\theta}$ such that reverse SDE can start from any point with a low value of $\gE$ and reach a mode of $\gE$. Thus, what remains is finding informative points to construct our \dem objective. 

\loose
To find informative points we start by first noting that \dem can be used as an off-policy objective, which means that the objective can be evaluated using any set of samples. Consequently, in problem settings where we have access to an initial dataset, \eg, from MCMC or MD simulations, we can readily leverage them to warm start training of $s_{\theta}$. In contrast, in settings with no initial samples, this feature is not possible. However, randomly exploring the sample space is unlikely to yield an informative $x_0$, especially in high dimensions where the energy landscape might be sparse.

\loose
We alleviate this cold-start problem by directly using our diffusion sampler $s_{\theta}$. 
In particular, we use the reverse SDE associated with $s_{\theta}$ which enables us to start from a mass covering prior, \eg, standard normal, and reach points that are progressively more informative samples. Note that we are free to choose a different diffusion coefficient $g(t) d\overline{w}_t$ in (\ref{eq:reverse_sde}) to increase or decrease the amount of exploration when generating a point $x_0$. In addition, we can run each reverse SDE in parallel to produce a batch of samples that we store in a replay buffer $\gB$. Buffer samples can then be used in the inner-loop; resembling a persistent contrastive-divergence objective~\citep{tieleman2008training} to train energy-based models. 

\loose
We highlight that in this outer loop sampling phase $s_{\theta}$ is fixed---\ie, the parameters $\theta$ are not updated---and as a result, despite simulation of the reverse SDE, \name is computationally cheap as we do not need to backpropagate gradients through the SDE solver. We note that making the algorithmic choice of sampling $x_0$ with the reverse $s_{\theta}$ \name can be viewed as a hybrid approach within the spectrum of on-policy to off-policy methods. This is due to the fact the forward SDE to get $x_t$ differs from the reverse SDE with a modified diffusion coefficient used to populate $x_0$ in $\gB$.

\loose
By training our diffusion sampler in every inner loop step we obtain an improved diffusion sampler. This in turn improves the fidelity of the batch of samples in the replay buffer $\gB$ produced by $s_{\theta}$. Thus, every pair of inner and outer loop operations in \name (see \cref{alg:idem}) produces a new sampler that is iteratively retrained and new sampled points that populate the buffer. From this perspective, we can view the process of learning as obtaining a higher-quality amortized sampler---mimicking a fully mixed MCMC algorithm---after the completion of the inner loop. Importantly, this sampler can be used in the \emph{absence} of any ground truth data and is the chief vehicle that allows \name to explore and find all salient modes of the energy function.

\subsection{Incorporating symmetries in \name}
\label{sec:equivariance}
\loose

Boltzmann-type distributions found in physical processes are beholden to the symmetries of the system. In this case, the symmetries arise from the spatial invariance of the energy function itself. More precisely, if we take $n$-body systems in $\R^d$, where $d=3 n$, the symmetries correspond to the rotation, translation, and permutation of the particles. These symmetries render the target density $\mu_{\text{target}}$ invariant to the product group $G = \sethree \times \mathbb{S}_n$. 

\loose
If $G$ is a subgroup of the orthogonal group $\text{O}(n)$---\ie, rotations and reflections---and carries an orthogonal action, then the gradient of a $G$-invariant function is $G$-equivariant~\citep[Lemma 2]{papamakarios2021normalizing}. As a result, we have that if the energy function $\E$ is $G$-invariant,  due to $\mu_{\text{target}}$ being $G$-invariant, the gradient $\nabla \E$ is $G$-equivariant. We can extend this result to the product group $\sethree \times \mathbb{S}_n$ by embedding this in $\text{O}(3n)$ by first projecting to a translation invariant subspace and defining an extended action\footnote{$\text{O}(3) \times \mathbb{S}_n$ action on $\R^{3n}$ is such that $\text{O}(3)$ acts diagonally while $\mathbb{S}_n$ acts by an orthogonal permutation matrix on the particles.} that acts orthogonally in $\R^{3n}$. 
Invoking the $\sethree \times \mathbb{S}_n$ symmetry constraint in the \dem objective leads to the following proposition:

\begin{propositionE}[][end,restate]
Let $G$ be the product group $\sethree \times \mathbb{S}_n \hookrightarrow \text{O}(3n)$ and $p_0$ be a $G$-invariant density in $\R^d$. Then the Monte Carlo score estimator of $\gS_K(x_t, t)$, is $G$-equivariant if the sampling distribution $x_{0\mid t} \sim \overline{\gN}(x_{0\mid t}; x_t, \sigma^2_t)$ is $G$-invariant, \ie,  $\overline{\gN}(x_{0\mid t}; g \circ x_t, \sigma^2_t) = \overline{\gN}(g^{-1} x_{0\mid t}; x_t, \sigma^2_t)$.
\label{prop:equivariance}
\end{propositionE}
\begin{proofE}
    \loose
   First note that since $p_0$ is $G$-invariant so, $\nabla p_0 (x_{0\mid t}^{(i)})$ is $G$-equivariant. This means $\gE$ and $\nabla \gE$ are $G$-invariant and $G$-equivariant respectively. A group element $g$ acts on $x \in \R^d$ in the standard way $g \circ x = gx$. $G$ acts on the space of distributions over $\R^d$: if $X$ is an $\R^d$-valued random variable with density $p$, then $g\circ X$ is distributed with density  $(g \circ p) (x) = p(g^{-1} x)$. Applying the group action to $\bar{\gN}$, we get
  \begin{align*}
    \gS_K(g \circ x_t,t) 
    & = \sum_i \frac{\exp(-\gE({g} \circ x_{0\mid t}^{(i)}))}{\sum_j\exp(-\gE({g} \circ x_{0\mid t}^{(j)}))} \nabla\gE({g}\circ x_{0\mid t}^{(i)}) \\
    & = {g}\circ \left(\sum_i \frac{\exp(-\gE(x_{0\mid t}^{(i)}))}{\sum_j\exp(-\gE( x_{0\mid t}^{(j)}))} \nabla\gE(x_{0\mid t}^{(i)}) \right) \\
    & = g\circ \gS_K(x_t,t),\\
x_{0\mid t}^{(1)},\dots,x_{0\mid t}^{(K)}&\sim\overline{\gN}(x_t,\sigma^2_t).
\end{align*}
Note that in the first line we used that $x_{0\mid t}^{(i)}\sim \overline{\gN}(x_t,\sigma^2_t)$ is equivalent to $g\circ x_{0\mid t}^{(i)}\sim \overline{\gN}(g\circ x_t,\sigma^2_t)$.
\end{proofE}
\loose
\vspace*{-10pt}
In practice, we can easily implement an equivariant $\gS_K(x_t,t)$ by replacing the standard normal distribution with a normal distribution that has zero center of mass. Note that a standard normal is already rotation and permutation invariant and an orthogonal action induces no change in volume as the determinant is $1$. Whilst it is not possible to define a translation invariant measure on $\R^d$ we can still achieve this symmetry by constructing a normal distribution that has zero center of mass. Intuitively, this is a projection of the density in $\R^d$ to $\R^{d-1}$ and this subspace is translation invariant~\citep{kohler2020equivariant,garcia2021n,midgley2023se}. Finally, to use symmetries within our \name algorithm we also need the diffusion sampler $s_{\theta}$ to be equivariant to the product group. While this design choice is necessary it does not come with any loss of generality as there always exists an equivariant map between two group invariant distributions on $\R^d$~\citep{bose2021equivariant}.

\section{Experimental results}
\label{sec:experiments}

\begin{figure*}[ht]
    \centering
    \includegraphics[width=1\linewidth]{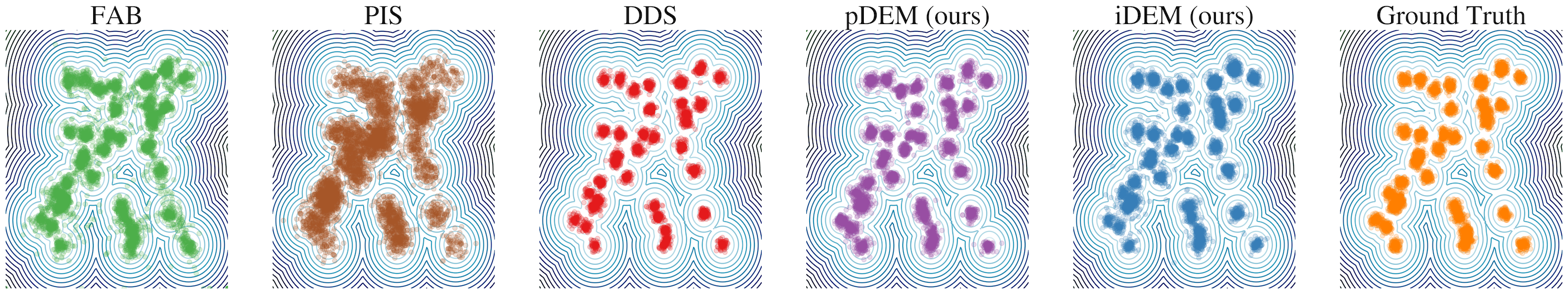}
    \caption{
         Contour lines for the target distribution, which is a GMM with $40$ modes. Colored points represent samples from each method.  
    } 
    \label{fig:gmm_figs}
\end{figure*}

\loose
We evaluate \name on multiple unnormalized densities including synthetic and $\sethree \times \mathbb{S}_n$-equivariant $n$-body particle systems of varying complexity as examples of scientific applications\footnote{Code for \name is available at \url{https://github.com/jarridrb/dem}.}. For our metrics, in \cref{tab:results_wide}, we report both sample-based metrics, such as 2-Wasserstein that assesses mode coverage, and Effective Sample Size (ESS), as well as the standard negative log-likelihood (NLL). We report additional metrics such as log partition function ($\log{Z}$) and Total Variation (TV) distance in \cref{tab:additional_metrics} in \S\ref{app:additional_metrics}

\loose
\xhdr{Datasets} We evaluate \name on four datasets, a $40$-Gaussian mixture model (GMM), and three equivariant potentials: A $4$-particle double-well potential (DW-4), a $13$-particle Lennard-Jones potential (LJ-13), and a $55$-particle Lennard-Jones potential (LJ-55) (see \autoref{app:experimental_setup:target_densities} for details). These benchmark datasets are chosen to demonstrate how scaling dimension $d$ affects algorithms, and due to their use in scientific applications~\citep{kohler2020equivariant,klein2023equivariant}. 

\xhdr{Baselines} We compare \name to three recent works: the path integral sampler (PIS)~\cite{zhang2021path}, denoising diffusion sampler (DDS)~\cite{vargas2023denoising}, and flow annealed bootstrapping (FAB)~\cite{midgley2022flow}. PIS and DDS are the most comparable models to \name as they are both diffusion-based but require simulating trajectories to evaluate their objective. On the other hand, FAB is the current state-of-the-art approach that combines AIS samples and (equivariant) normalizing flow training within a buffer. We also include prior-\dem (\pdem) which fills the buffer with the $\sethree \times \mathbb{S}_n$-invariant prior.

\loose
\xhdr{Architecture} For \name, PIS, and DDS, we can use any network $s_\theta: ( \R^d, \R^+) \to \R^d$. We use an MLP with sinusoidal positional embeddings for the GMM and an EGNN flow model architecture~\citep{satorras2021n} for the equivariant densities (DW-4, LJ-13, and LJ-55) following~\citet{klein2023equivariant}. FAB, however, requires a specialized invertible architecture, so we use the architecture from \citet{midgley2022flow} for GMM and $\sethree$-augmented coupling flow architecture from \citet{midgley2023se} for the equivariant tasks. Finally, in our parametrization of \name, we sometimes find it useful to pin the score at $t=0$ to $\nabla \log p_0(x_0)$ as we have access to it and do not need to estimate it with MC---\ie, $\gS_K(x_0, 0)$. We provide further details on the experimental setup in \autoref{app:experimental_setup:architecture}.

\subsection{Main results}
\label{sec:main_results}

\begin{table*}[t]
\caption{\small Sampler performance with mean $\pm$ standard deviation over 3 seeds for negative log-likelihood (NLL), Effective Sample Size (ESS), and 2-Wasserstein metrics ($\mathcal{W}_2$). $*$ indicates divergent training. \textbf{Bold} via Welch's two sample t-test $p < 0.1$. See \S\ref{app:experimental_setup:results_wide} for more details.}
\label{tab:results_wide}
\resizebox{1\linewidth}{!}{
\begin{tabular}{@{}lcccccccccccc}
    \toprule
    Energy $\rightarrow$ & \multicolumn3c{GMM ($d=2$)} & \multicolumn3c{DW-4 ($d=8$)} & \multicolumn3c{LJ-13 ($d=39$)} & \multicolumn3c{LJ-55 ($d=165$)} \\
    \cmidrule(lr){2-4}\cmidrule(lr){5-7}\cmidrule(lr){8-10}\cmidrule(lr){11-13}
    Algorithm $\downarrow$ & NLL & ESS & $\gW_2$ & NLL & ESS & $\gW_2$ & NLL & ESS & $\gW_2$ & NLL & ESS & $\gW_2$ \\
    \midrule
    FAB~\citep{midgley2022flow} 
    & 7.14\std{0.01}             & 0.653 \std{0.017}  & 12.0\std{5.73} 
    & \textbf{7.16\std{0.01}}    &\textbf{0.947 \std{0.007}} & 2.15\std{0.02} 
    & \textbf{17.52\std{0.17}}   &0.101 \std{0.059} & 4.35\std{0.01} 
    & 200.32\std{62.3}          &0.063 \std{0.001} & 18.03\std{1.21} \\
    PIS~\citep{zhang2021path}
    & 7.72\std{0.03}   & 0.295 \std{0.018} & \textbf{7.64\std{0.92}}
    & 7.19\std{0.01}   & 0.901 \std{0.003} & \textbf{2.13\std{0.02}}
    & 47.05\std{12.46} & 0.004 \std{0.002} & 4.67\std{0.11}
    & $*$            & $*$ & $*$ \\
    DDS~\citep{vargas2023denoising}
    &7.43\std{0.46}  & 0.687 \std{0.208} & 9.31\std{0.82}
    &11.27\std{1.24} & 0.408 \std{0.001} & 2.15\std{0.04}
    & $*$            & $*$ & $*$
    & $*$            & $*$ & $*$ \\
    \pdem (ours)
    & 7.10\std{0.02} & 0.634 \std{0.084}  & 12.20\std{0.14}
    & 7.44\std{0.05} &0.547 \std{0.010} & \textbf{2.11\std{0.03}}
    & 18.80\std{0.48} & 0.044 \std{0.013} & \textbf{4.21\std{0.06}}
    & $*$            & $*$ & $*$ \\
    \name (ours) 
    & \textbf{6.96\std{0.07}}    & \textbf{0.734 \std{0.092}} & \textbf{7.42\std{3.44}}
    & \textbf{7.17\std{0.00}}    & 0.825 \std{0.002} & \textbf{2.13\std{0.04}} 
    & \textbf{17.68\std{0.14}}   & \textbf{0.231 \std{0.005}} & \textbf{4.26\std{0.03}}
    & \textbf{125.86\std{18.03}}  & \textbf{0.106 \std{0.022}} & \textbf{16.128\std{0.071}} \\

    \bottomrule
    \end{tabular}
}

\end{table*}

\loose
We report the sample likelihood-based metrics in \autoref{tab:results_wide}. For a fair comparison between \name and all baselines---some of which cannot readily provide an NLL---we fit an optimal transport conditional flow matching (OT-CFM) model~\citep{tong_conditional_2023} on generated samples from each method. We use the reverse ODE of this OT-CFM model to compute a test NLL which is presented in \autoref{tab:results_wide}. We find that \name outperforms all considered baselines on $\gW^2_2$ and TV indicating high-quality generated samples. This result is also qualitatively substantiated in \autoref{fig:gmm_figs} for GMMs where we notice \name and DDS obtain the best samples. 

\loose
For NLL we find that \name matches or outperforms all baselines on GMM, DW-4, and LJ-13. Importantly, on the most challenging and high-dimensional energy LJ-55, unlike \name which obtains the best NLL, PIS and DDS experience unstable training and cannot learn successfully on the task. Thus, 
 We reconcile this by noting that LJ-55 has an energy with a high Lipschitz constant (see \autoref{ref:app_lj}) and without smoothing represents a significant modeling challenge.

\begin{figure}[t]
    \captionsetup[subfigure]{aboveskip=-1pt,belowskip=-1pt}
    \begin{subfigure}{0.49\linewidth}
        \centering
        \includegraphics[width=\linewidth]{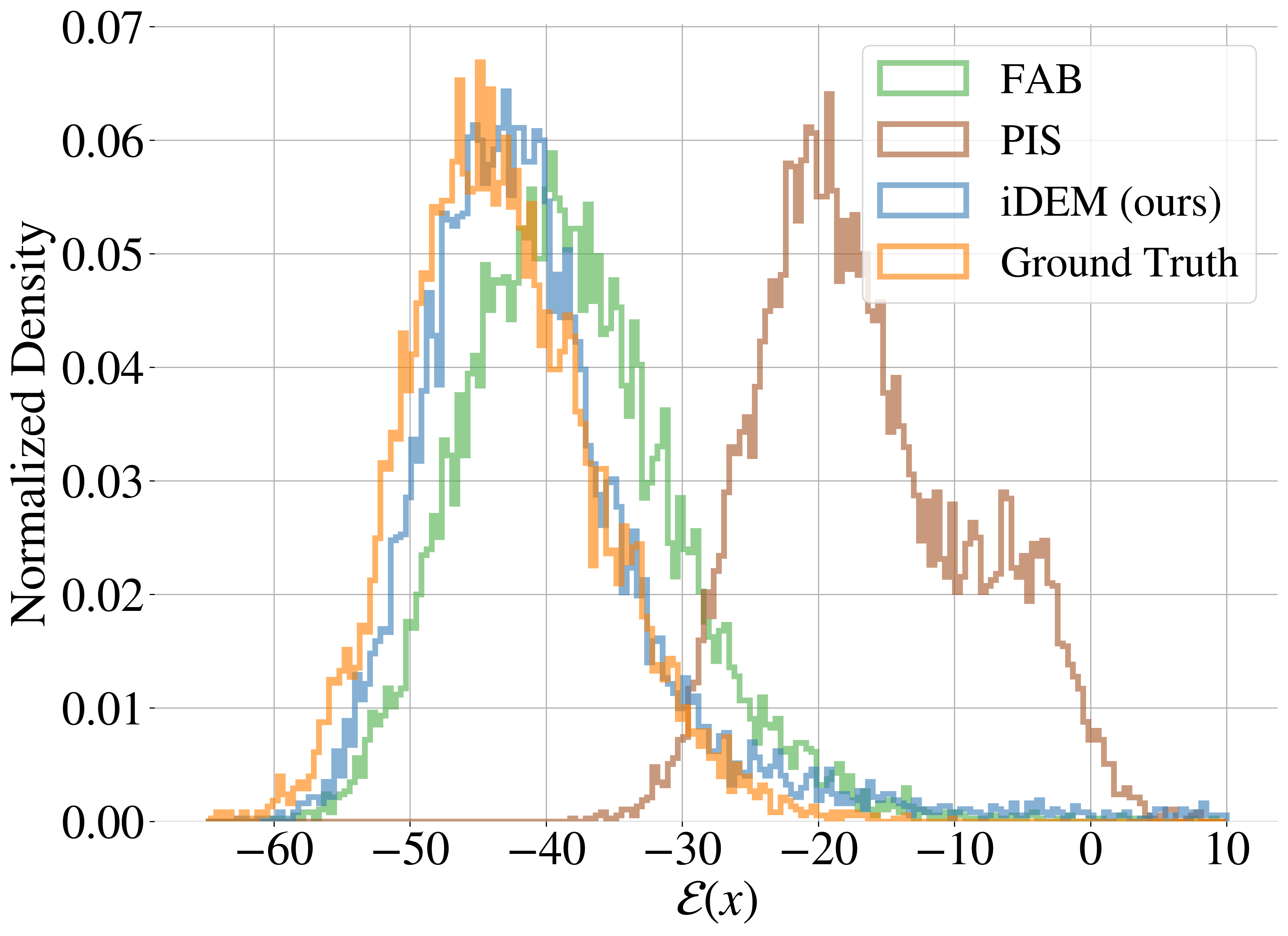}
        \label{fig:lj13_energy}
    \end{subfigure}
    \begin{subfigure}{.49\linewidth}
    \centering
        \includegraphics[width=\linewidth]{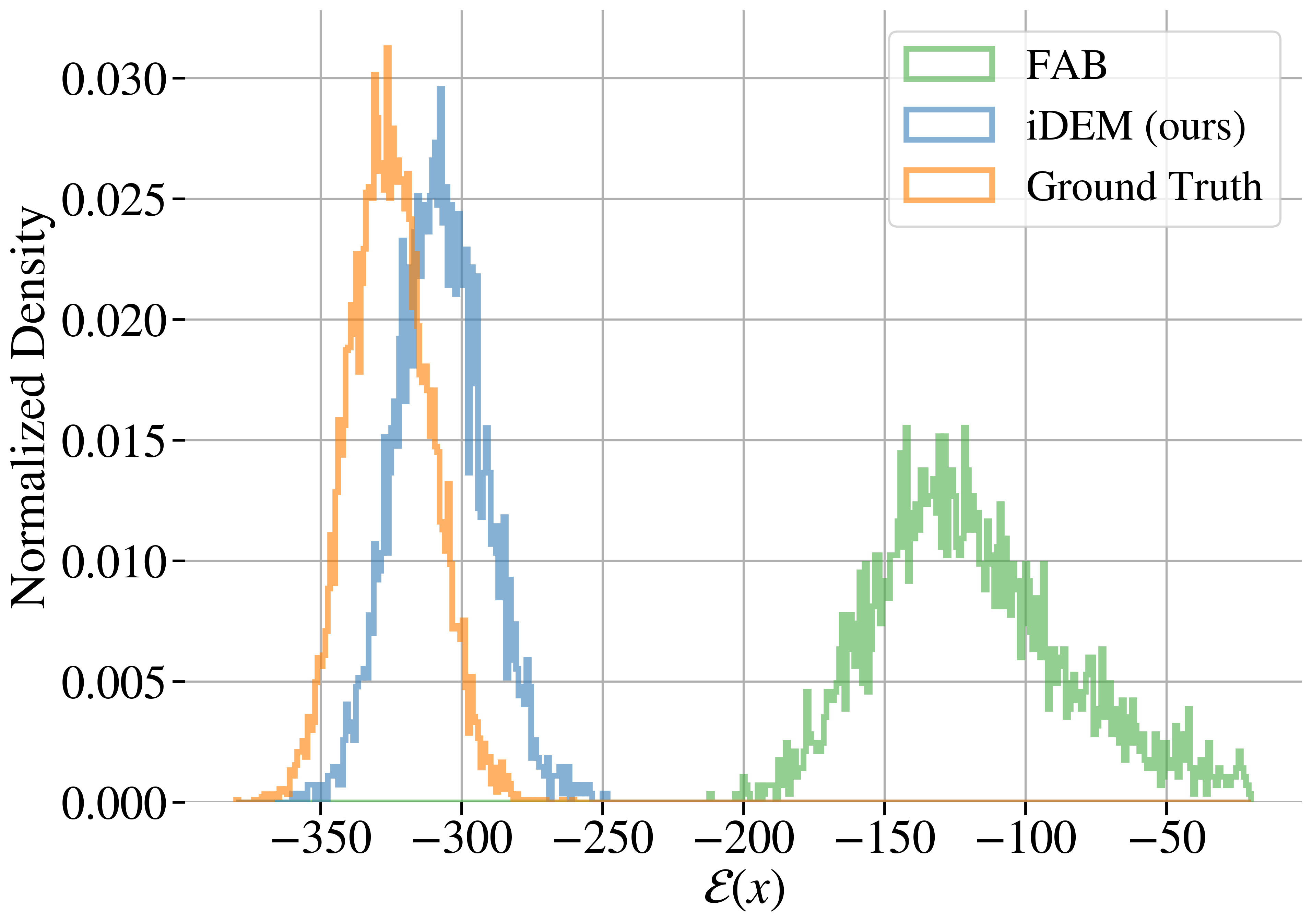}
        \label{fig:lj55_energy}
    \end{subfigure}
    \vspace{-20pt}
    \caption{ \small
        Comparison of the ground truth energy histograms of LJ-13 (left) and LJ-55 (right) and energies of samples generated from various methods. DDS is omitted from both plots while PIS is omitted from LJ-55 as they diverge in these settings.
    }
    \label{fig:main_energy_figs}
\end{figure}

In \autoref{fig:main_energy_figs} we visualize the energy histograms of the LJ-13 and LJ-55 systems (see \autoref{app:additional_results} for DW-4) in comparison to model samples. We also report inter-atomic distances for samples from the training data and all models in \autoref{app:interatomic_distance}. We observe that \name is the best approach in terms of accurately modeling the true energy of each system with a significant separation between \name and FAB, the second best approach, on the LJ-55 energy. Notably, methods such as DDS and PIS are unable to train properly on this task.

\loose
\xhdr{Computational complexity}
We quantify the computational footprint of each method by reporting training time in hours to convergence in \autoref{tab:train_time}. We find that \name is significantly faster than the previous SOTA FAB on all tasks due to FAB being bottlenecked by AIS. In particular, \name is $\mathbf{\sim 4\times}$ faster on high dimensional tasks LJ-13 and LJ-55 while $\mathbf{\sim 1.8\times}$ faster on the lower dimensional, less computationally expensive GMM and DW-4 tasks. Furthermore, \name is also faster than neural sampler baselines like PIS and DDS, which we attribute to the simulation-free gradients in our \dem objective.  Finally, we observe that training times for \pdem are significantly smaller than all other methods due to it being truly simulation-free. However, this comes at the cost of training stability---2 of 3 \pdem runs diverged on LJ-55.

\begin{table}[t]
    \centering
    \caption{Training time results in hours excluding evaluation time. $*$ denotes divergent training runs.}
\resizebox{\columnwidth}{!}{
    \begin{tabular}{@{}l c c c c}
    \toprule
    Algorithm $\downarrow$ Dataset $\rightarrow$ & GMM & DW-4 & LJ-13 & LJ-55\\
    \midrule
    FAB~\citep{midgley2022flow} & 1.71 & 6.87 & 21.78 & 40.35 \\ 
    PIS~\citep{zhang2021path} & 4.11 & 11.29 & 17.36 & $*$ \\ %
    DDS~\citep{vargas2023denoising} & 1.81 & 5.65 & $*$ & $*$\\ %
    \pdem (ours) & 0.36 & 1.40 & 1.79 & $*$ \\ %
    \name (ours) & 0.87 & 4.30 & 6.55 & 7.75\\ 
    \bottomrule
    \end{tabular}
    }
    \label{tab:train_time}
    \vspace{-10pt}
\end{table}

\subsection{Ablation experiments}
\label{sec:ablation_experiments}
\loose
We next investigate different aspects of the \name in a set of ablation studies that seek to answer a series of questions (\textbf{Q1}-\textbf{Q3}) using the GMM, DW-4, and LJ-13 energies. We also include additional ablation experiments in \autoref{app:further_ablations}.

\begin{figure}[t]
    \captionsetup[subfigure]{aboveskip=-1pt,belowskip=-1pt}
    \begin{subfigure}{0.49\linewidth}
        \centering
        \includegraphics[width=\linewidth]{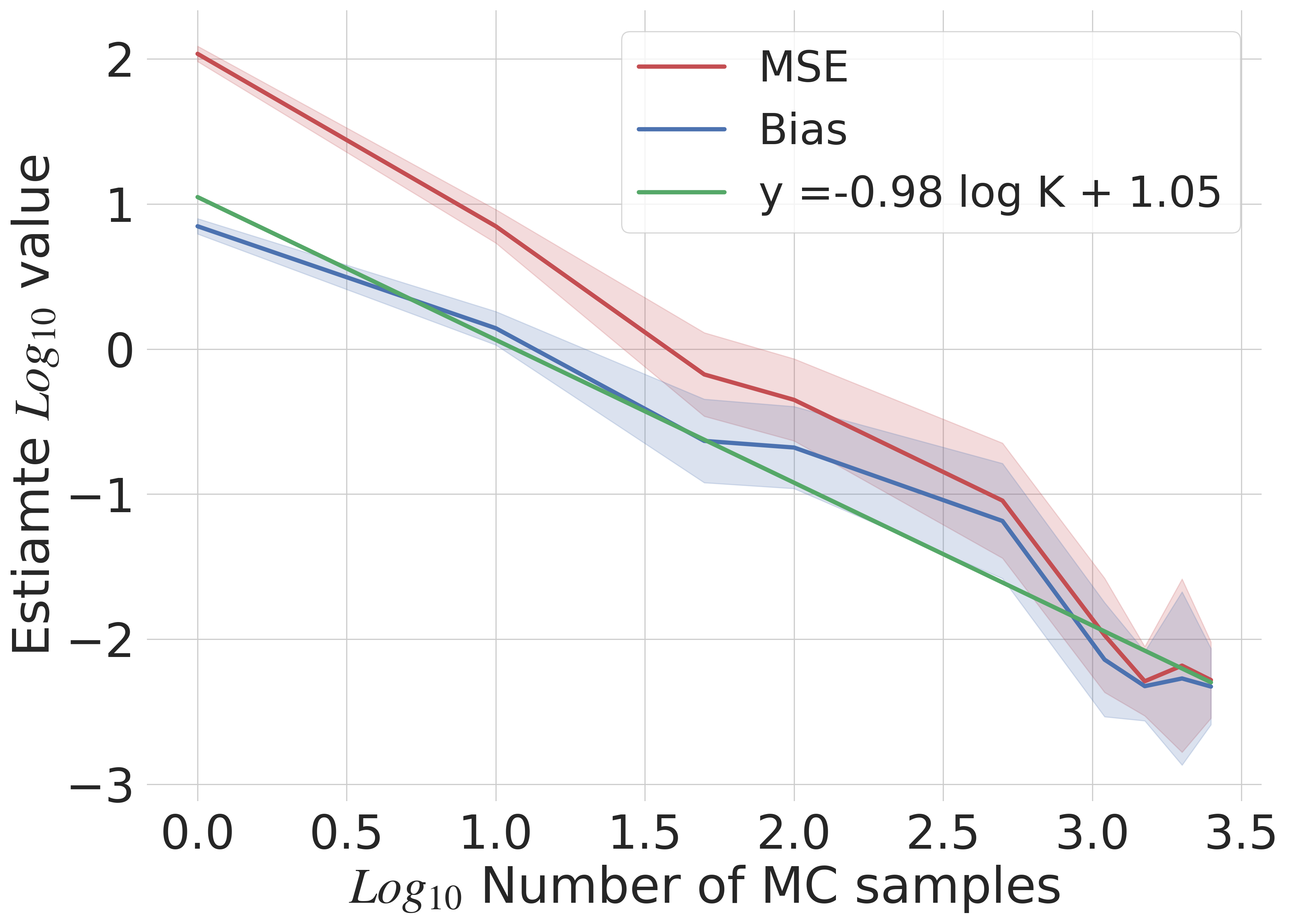}
        \label{fig:ablation_fig:a}
    \end{subfigure}
    \begin{subfigure}{.49\linewidth}
    \centering
        \includegraphics[width=\linewidth]{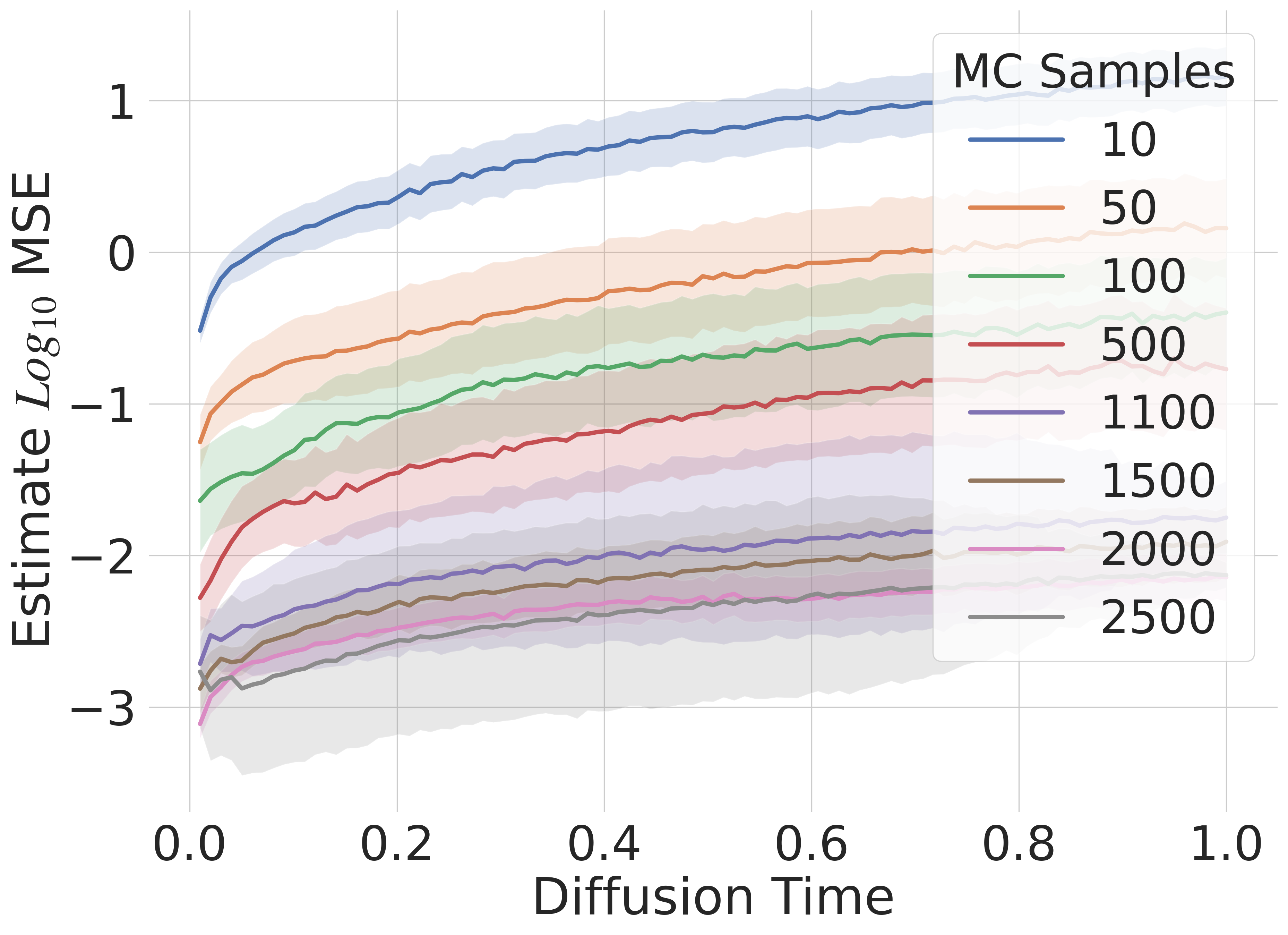}
        \label{fig:ablation_fig:b}
    \end{subfigure}
    \vspace{-15pt}
    \caption{\small \textbf{Left:} Log-log plot of bias and MSE vs.\ $K$ and a regression to the bias. \textbf{Right:} Plot of log bias vs.\ energy for different $K$. The MSE and bias are calculated for GMM with a linear noise schedule. The standard deviations for the log-transformed values are over $10$ seeds with the variance estimated over $256$ samples. For the plot on the right, the values are averaged over $x_0 \sim p_0$.}%
    \vspace{-6pt}
    \label{fig:ablation_fig}
\end{figure}

\loose
\xhdr{Q1: Bias and MSE of \dem vs.\ $K$}
\loose
In \autoref{fig:ablation_fig} (left) we report the bias and mean squared error (MSE) of $\gS_K$ versus $K$ on the GMM in log-log plot.
We find that as $K \to \infty$ the bias and MSE decrease as we increase $K$ and in particular the bias goes to $0$ which verifies that \dem is a consistent estimator. Additionally, a linear regression to the bias reveals an asymptotic decay rate of $O(1/K)$, which empirically validates \autoref{prop:bias} with a slightly sharper rate than $O(1/\sqrt{K})$.

\loose
\xhdr{Q2: MSE of \dem vs. $t$  for different $K$}
\loose
In \autoref{fig:ablation_fig} (right) we study the log-MSE as a function of diffusion time---\ie, $x_t$ for $t \in [0,1]$---versus $K$ on GMM. As observed, the log MSE drops as we increase the number of MC samples but increases as $t \to 1$ as we get closer to the prior. As we increase time the diffusion process moves us farther from the modes of $\gE$, which means that our estimator has a higher bias for the same $K$. This is also empirically observed and supports the finding in the ablation experiments in the main paper \autoref{sec:ablation_experiments} and is a consequence of \autoref{prop:bias}.

\begin{figure}[t]
    \captionsetup[subfigure]{aboveskip=-1pt,belowskip=-1pt}
    \begin{subfigure}{0.49\linewidth}
        \centering
        \includegraphics[width=\linewidth]{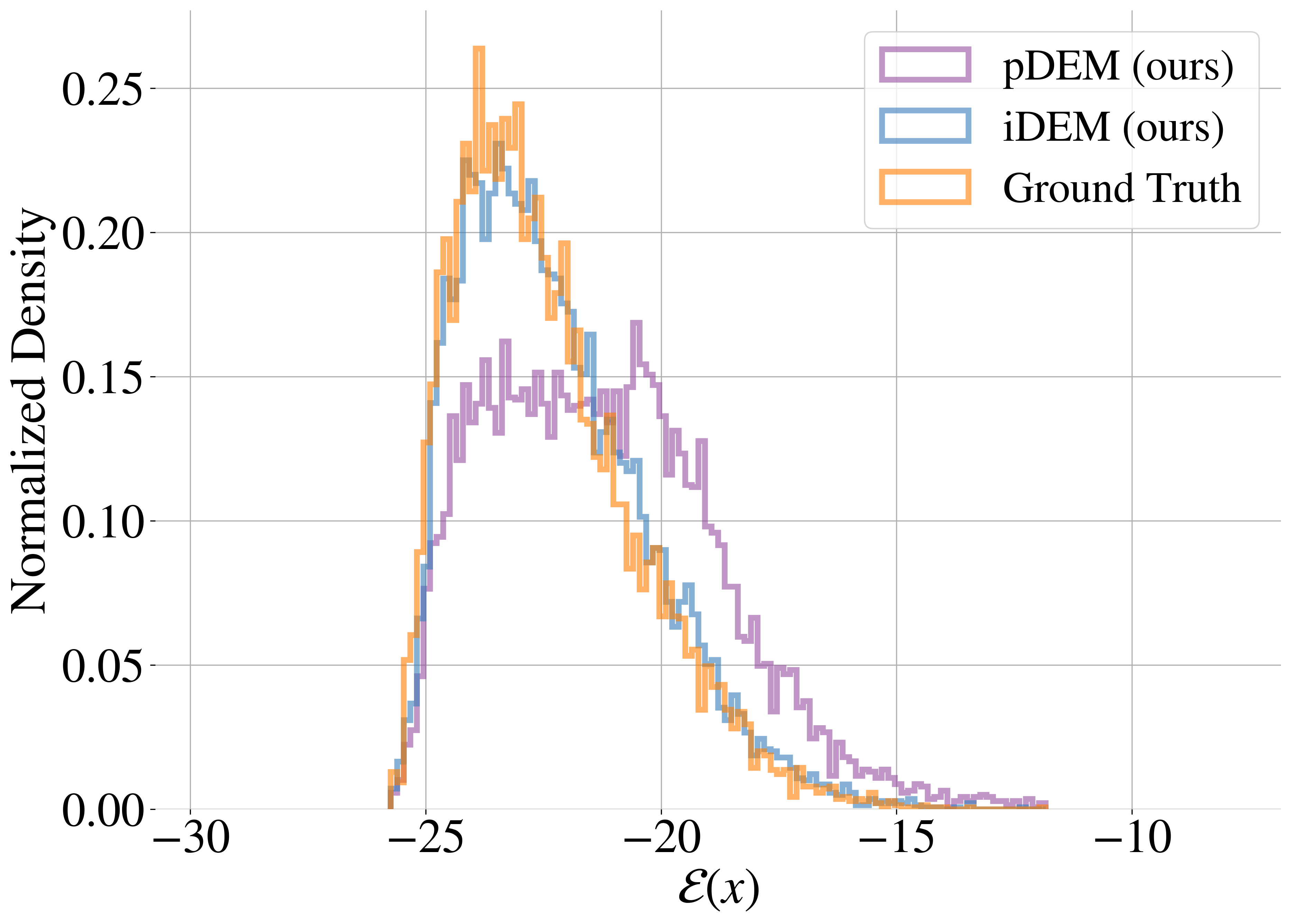}
        \label{fig:buffer_fig:a}
    \end{subfigure}
    \begin{subfigure}{.49\linewidth}
    \centering
        \includegraphics[width=\linewidth]{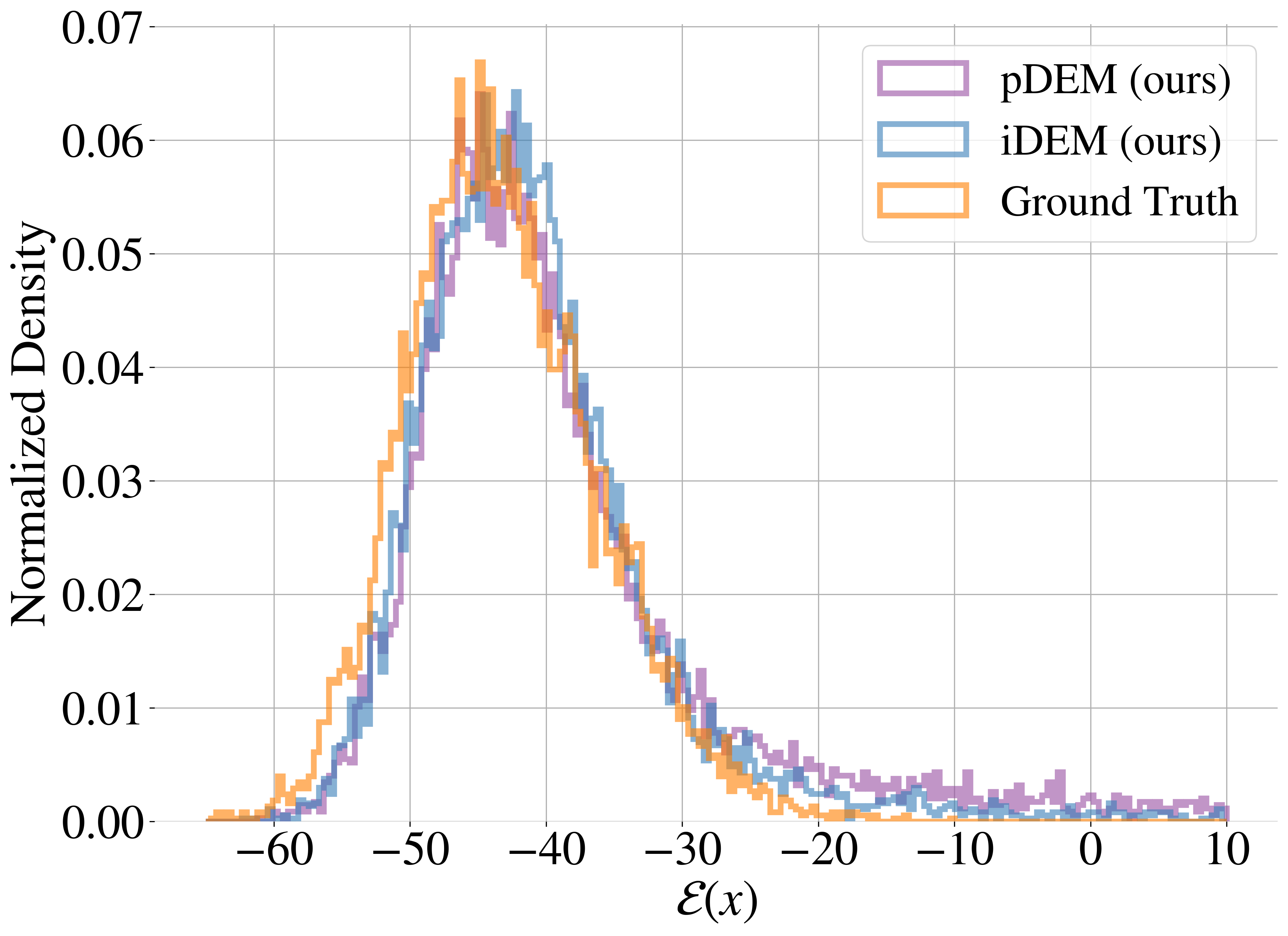}
        \label{fig:bufferfig:b}
    \end{subfigure}
    \vspace{-15pt}
    \caption{ \small
        Comparison of the ground truth energy histograms of DW-4 (\textbf{left}) and LJ-13 (\textbf{right}) in relation to energies of samples generated from \pdem and \name.
    }
    \vspace{-15pt}
    \label{fig:buffer_fig}
\end{figure}

\xhdr{Q3: The utility of a buffer in \name}
\loose
We study the performance of \dem with and without samples from $s_{\theta}$ in a buffer. In particular, we 
ablate \name to prior-\dem (\pdem), which is a pure off-policy method, requiring no simulation-based outer loop. We see a clear trend on \autoref{fig:buffer_fig} (left) for DW-4 energy histograms, where using samples from $s_{\theta}$ leads to better performance. On LJ-13 plotted on \autoref{fig:buffer_fig} (right) \name and \pdem perform roughly the same and there are minor improvements in using $s_{\theta}$. Finally, on LJ-55 \pdem failed to learn in $2$ out of $3$ runs, which highlights the increased stability of \name over \pdem. %

\section{Related work}
\label{sec:related_work}

\loose
\xhdr{MCMC and variational approximations}
MC methods like AIS~\citep{neal2001annealed} and SMC~\citep{del2006sequential} are often regarded as gold standards for sampling but are expensive and often hampered by slow convergence~\citep{robert1999monte}. Variational techniques such as mean-field approximation~\citep{wainwright2008graphical} and amortized methods like normalizing flows~\citep{papamakarios2021normalizing} are appealing alternatives for distribution approximation. Hybrid approaches that combine flows and MCMC to improve the transition kernels~\citep{wu2020stochastic,geffner2021mcmc,thin2021monte,doucet2022score,geffner2023langevin,grenioux2023sampling} are an attractive compromise and have shown empirical benefits, \eg, FAB~\citep{midgley2022flow}, CRAFT~\citep{matthews2022continual}. Similar in approach to \name,  \citet{song2023loss} applies Monte Carlo approximation to the guidance term for solving inverse problems with diffusion models.

\loose
\xhdr{Equivariant flows and Boltzmann generators}
Several key works use Boltzmann generators to sample from unnormalized probability densities~\citep{noe2019boltzmann}. These include equivariant approaches using normalizing flows~\citep{kohler2020equivariant,midgley2023se,klein2023equivariant,kohler2023rigid}. MD simulations have also seen the benefits of flow-based proposal distributions~\citep{klein2023timewarp}. Generative models for $\sethree$-equivariant distributions span application domains such as robotics~\citep{brehmer2023edgi,brehmer2023geometric}, molecular modeling~\citep{hoogeboom2022equivariant,xu2022geodiff,igashov2022equivariant}, and protein generation~\citep{yim2023se,yim2023fast,bose2023se}.

\looseness=-1
\xhdr{Neural samplers} Motivated by the Schr\"odinger bridge problem as a unifying perspective linking generative modeling to stochastic control~\citep{pavon1989stochastic,dai1991stochastic,tzen2019theoretical}, neural samplers seek to amortize MCMC.
Most similar to our approach are the works of~\citet{berner2022optimal,vargas2023denoising,zhang2021path,richter2023improved,vargas2024transport} which exploit diffusion processes for fast mode mixing. However, these approaches require simulation to compute the objective, unlike \name. Finally, \name uses an iterative scheme where the sampler is trained on modifications of its own initial samples, resembling training diffusion models on their on data~\citep{bertrand2023stability,alemohammad2023self}.

\looseness=-1
\xhdr{GFlowNets} Continuous generative flow networks \citep{lahlou2023cgfn} are deep reinforcement learning algorithms that have the explicit aim of \emph{off-policy} training of sequential samplers, diffusion-structured samplers being a particular case (\citet{zhang2023unifying,lahlou2023cgfn}, \S4.2). These methods can stably learn from sampled states or trajectories without differentiation through the simulated process that produced them \citep{malkin2022gfnhvi}. Avoiding SDE integration in the training loop is one of the motivations for our work, and \name can be seen as a simulation-free training algorithm for generative flow networks of a certain structure. 

\vspace{-2pt}
\section{Conclusion}
\label{sec:conclusion}

\loose
In this paper, we tackle the problem of amortized sampling from Boltzmann distributions. Our proposed \name algorithm uses a novel stochastic matching loss in the inner loop to train a diffusion sampler. Exploiting the amortization benefits of the diffusion sampler, we leverage it to propose informative samples to further accelerate its training. Empirically, we find \name to be significantly faster than previous approaches while offering high mode coverage and state-of-the-art performance on multiple benchmarks and is the first approach that is scalable to the challenging LJ-55 for energy-based training. While \name is computationally cheap, the \dem objective is biased and may be affected by the variance of the samples. Reducing the variance of \dem, including with adaptive techniques~\citep{bugallo2017adaptive},  and leveraging advances in SDE simulation to speed up the outer loop are natural directions for future investigation. %

\section*{Impact statement}
\label{sec:impact_statement}
\looseness=-1
This work studies amortized sampling from Boltzmann densities, a problem of general interest in machine learning that arises both in pure statistical modeling (\eg, sampling high-dimensional Bayesian posteriors over parameters) and in applications. We highlight the molecular design task---in turn applicable to drug and material discovery---as a motivation and immediate application for \name. While we do not foresee immediate negative impacts of our advances in this area, we encourage due caution to prevent their potential misuse.

\section*{Contribution statement}

J.R., N.M., and A.T. initially conceived the idea of a stochastic off-policy continuous regression objective. A.J.B., T.A., G.G., and A.T. independently approached the problem through the lens of active inference with a generative model in a bi-level iterative scheme. 
Experiments were primarily led by T.A., J.R., A.T. (\name) and J.R,S.M., P.L., C.H.L., M.S. (ablations and baselines).
G.G., A.J.B., and N.M. led the development of the theory. 
S.R., G.G., and Y.B. guided the project. A.J.B. and N.M. drove the writing of the paper, with contributions from all other authors.
All authors contributed to designing the experiments.

\section*{Acknowledgments}
The authors would like to thank James Vuckovic, Raymond Chua, Karam Ghanem, and Christos Tsirigotis for useful comments on early versions of this manuscript. In addition, the authors thank Julius Berner for sharing their code for PIS and DDS. A.J.B. is supported through an NSERC Post-doctoral fellowship. 

The authors acknowledge funding from UNIQUE, CIFAR, NSERC, Intel, and Samsung. The research was enabled in part by computational resources provided by the Digital Research Alliance of Canada (\url{https://alliancecan.ca}), Mila (\url{https://mila.quebec}), Dreamfold, Anyscale, Google GCP, and NVIDIA.

\clearpage
\bibliography{clean}
\bibliographystyle{style/icml2024}

\appendix
\onecolumn

\section{Proofs of propositions}
\label{sec:proofs}

\printProofs

\section{\name for non-VE noising processes}
\label{app:non_vesde}

We sketch how the objective described in \autoref{sec:denoising_boltzmann_target} can be generalized to general noising processes of the form (\ref{eq:ou}).

Consider a SDE of the form $dx_t=-\alpha(t)x_t\,dt+g(t)\,dw_t$, and define
\begin{equation}
    y_t := \beta(t)x_t,\quad\beta(t):=\exp\left(-\int_0^t\alpha(s)\,ds\right).
    \label{eq:ito_transformation}
\end{equation}
We have $y_0=x_0$, and, by It\^o's lemma, $y_t$ also obeys a SDE:
\begin{align}
    dy_t
    &=[\beta'(t)x+\beta(t)\alpha(t)x]dt+g(t)\beta(t)\,dw_t\nonumber\\
    &=g(t)\beta(t)\,dw_t,
    \label{eq:ito_to_ve}
\end{align}
where we have used that $\beta'(t)=-\beta(t)\alpha(t)$ by the definition (\ref{eq:ito_transformation}). The SDE (\ref{eq:ito_to_ve}) is variance-exploding, and the analysis in \autoref{sec:denoising_diffusion} (for diffusion models) or in \autoref{sec:denoising_boltzmann_target} (for \name) applies to $y_t$. 

This VE SDE generates marginal densities $\tilde p_t(y_t)$ from the initial distribution $\tilde p_0=p_0$. An estimator of the score $\nabla\log\tilde p_t(y_t)$, which can be fit using the mentioned algorithms, is equivalent to an estimator of $\nabla\log p_t(x_t)$, since  $\nabla\log\tilde p_t(y_t)=\nabla\log p_t(x_t)$. (Whether the neural network estimator takes $x_t$ or the rescaled $y_t$ as input is an implementation choice; we use $x_t$ as input for numerical stability.)

Finally, we note that the above is readily generalized to the case of noising SDEs with matrix coefficients, $dx_t=A(t)x\,dt+G(t)\,dw_t$, a case which may be of interest in future generalizations of \name to Lie group-equivariant settings.

\section{\name and flow matching}
\label{app:fm}

We remarked in \autoref{sec:denoising_boltzmann_target} that the denoising diffusion objective (\ref{eq:score_as_expectation}) and the \name regression target (\ref{eq:score_as_convolution}) express the score of the convolution in different ways, with the diffusion objective using
\[\nabla p_t = \nabla(p_0*\gN(0,\sigma_t^2)) = p_0*\nabla\gN(0,\sigma_t^2)\]
and the \name objective using
\[\nabla p_t = \nabla(p_0*\gN(0,\sigma_t^2)) = \nabla p_0*\gN(0,\sigma_t^2).\]
It is interesting to consider the former expression in the case of a Boltzmann target distribution. We have, as in (\ref{eq:score_as_expectation}),
\begin{equation}
    \nabla\log p_t(x_t)=\sE_{x\sim p_0(x)\gN(x;x_t,\sigma_t^2)}[\nabla\gN(x;x_t,\sigma_t^2)]=\sE_{x\sim p_0(x)\gN(x;x_t,\sigma_t^2)}\left[\frac{x-x_t}{\sigma_t^2}\right].
    \label{eq:score_as_other_convolution_boltzmann}
\end{equation}
The quantity inside the expectation is the appropriately scaled velocity of a line segment from $x_t$ to $x$. The expectation is intractable to compute, and \citet{huang2023reverse} recently proposed to estimate this expectation using Monte Carlo samples at generation time and showed bounds on the variance of the estimate. This can also be understood as a simulation-free estimation of the Föllmer drift; see \citet{zhang2021path}, \S3.1.

This expression also recalls the method of \emph{flow matching} \citep{lipman_flow_2022} for fitting the probability flow ODE of a stochastic process---a continuous normalizing flow (CNF)---given its boundary marginals.\footnote{Note that for a VE SDE, the probability flow ODE is simply $dx_t=\frac{-g(t)^2}{2}\nabla\log p_t(x_t)\,dt$, so fitting the ODE amounts to learning the score.} Generalizations of flow matching by \cite{liu_rectified_2022,albergo_building_2023,tong_conditional_2023} allow learning ODEs linking arbitrary marginal distributions and amount to regressing the ODE's drift at $x_t$ to the expected velocity of a line segment linking a source point to a target point and passing through $x_t$. 

The expression (\ref{eq:score_as_other_convolution_boltzmann}) is the target of flow matching for a certain interpolant density, but the expectation over $p_0(x)\gN(x;x_t,\sigma_t^2)$ is intractable. \citet{tong_conditional_2023} proposed an importance weighting solution that trains the CNF using flow matching on a weighted dataset of points $x_0$, allowing approximate flow matching with Boltzmann target densities. However, the results of this paper lead us to speculate about variants that estimate the regression target at $x_t$ directly, as \name does, and take advantage of a buffer of past generated points $x_0$ for efficient simulation-free training.

\section{Sampling with MCMC}
\label{app:sampling_app}

This section focuses on the most popular methods for sampling from complex distributions that do not rely on deep learning. We provide a brief summary of some existing methods, but we refer the reader to the corresponding literature for more details.

\subsection{Metropolis-Hastings}
\label{app:mh}

Perhaps the most commonly used sampling algorithm is the Metropolis-Hastings (MH) algorithm. The MH algorithm is a class of MCMC algorithm~\citep{metropolis1953equation,hastings1970monte}, meaning based on the idea of constructing a Markov chain whose stationary distribution is the target distribution $\mu_{\text{target}}$. 

The algorithm starts by sampling an initial state $x_0$ from some proposed initial distribution $q(x_0)$, and then iteratively samples a new state $x_{t+1}$ from a proposal distribution $q(x_{t+1} \mid x_t)$, typically a Gaussian distribution centred at $x_t$. The new state is accepted with probability $\alpha(x_t, x_{t+1})$, where

\begin{equation}
    \label{eq:alpha}
    \alpha(x_t, x_{t+1}) = \min\left\{1, \frac{p_{\text{target}}(x_{t+1})q(x_t \mid x_{t+1})}{p_{\text{target}}(x_t)q(x_{t+1} \mid x_t)}\right\}.
\end{equation}

The MH algorithm is guaranteed to converge to the target distribution if the detailed balance condition is satisfied:

\begin{equation}
    p_{\text{target}}(x_t)q(x_{t+1} \mid x_t) = p_{\text{target}}(x_{t+1})q(x_t \mid x_{t+1}).
\end{equation}
However, the algorithm can be very slow to converge, especially in high dimensions, and it is often necessary to use annealing schemes (\autoref{app:ais}) to improve mode coverage.

\subsection{Hamiltonian Monte Carlo}
\label{app:hmc}

Hamiltonian Monte Carlo~\cite{neal2011mcmc} is a different type of MCMC algorithm that uses the gradient of the energy function to construct a Markov chain that converges to the target distribution $\mu_{\text{target}}$. Similarly to MH, the algorithm starts by sampling an initial state $x_0$ from some proposed initial distribution $q(x_0)$. It then proceeds to iteratively sample new states $x_{t+1}$ by simulating the Hamiltonian dynamics of a particle with mass $m$ and position $x_t$ in a potential energy field $-\E(x) = \log \mu_{\text{target}}$, for a fixed time $T$. Hamiltonian dynamics are given by the following system of differential equations:

\begin{equation}
    \begin{aligned}
        \frac{dx}{dt} &= \frac{p}{m} \\
        \frac{dp}{dt} &= -\nabla \E(x),
    \end{aligned}
\end{equation}

The HMC algorithm samples an auxiliary new momentum $p_{t+1}$ from a Gaussian distribution centred at $0$ and then uses an integrator, such as the leapfrog integrator, to simulate the Hamiltonian dynamics for a fixed time $T$. The new state $x_{t+1}$ is then accepted with probability $\alpha(x_t, x_{t+1})$, given by (\ref{eq:alpha}). The HMC algorithm is more efficient than MH, thanks to the use of the gradient of the energy function, but it still struggles with mode coverage. Furthermore, the performance of the algorithm is very sensitive to tuning of the step size $T$ and the mass $m$, although automatic tuning methods have been proposed, such as the No U-Turn Sampler~\citep{hoffman2014no}.

\subsection{Metropolis-Adjusted Langevin Algorithm}

An alternative algorithm that uses the gradient of the energy function is the Metropolis Adjusted Langevin Algorithm (MALA)~\citep{grenander1994representations, roberts1996exponential, roberts1998optimal}. Similarly to HMC, the algorithm starts by sampling an initial state $x_0$ from some proposed initial distribution $q(x_0)$. It then proceeds to iteratively sample new states $x_{t+1}$ by simulating the Langevin dynamics of a particle with position $x_t$ in a potential energy field $-\E(x) = \log \mu_{\text{target}}$, for a fixed time $T$. Langevin dynamics are given by the following system of differential equations:

\begin{equation}
    \frac{dx}{dt} = -\nabla \E(x) + \sqrt{2} \xi,
\end{equation}
where $\xi$ is a standard Gaussian noise. 

\subsection{Annealed importance sampling}
\label{app:ais}

Markov chains have bad mode coverage, as the probability of jumping between separate modes can be extremely low. 
To help discover modes better we can combine chains with an annealing scheme where intermediate distributions $p_j$ and $p_{j+1}$ only differ slightly. For instance, in simulated annealing~\citep{kirkpatrick1983optimization}, we can choose $n$ distributions such that $p_n$ allows for high mode coverage by
setting $p_j \propto \mu_{\text{target}}^{\beta_j}$, for $1 =  \beta_0 > \beta_1 > \dots > \beta_L \geq 0$.

By viewing the annealing process as an importance sampling distribution we can derive a new estimator that has reduced variance in comparison to IS while achieving high mode coverage due to annealing. The Annealed Importance Sampling approach, like simulated annealing, considers a sequence of distributions, $\log p_j (x) = \beta_j \mu_{\text{target}}(x) + (1-\beta_j) \log p_L(x)$, where the intermediate samples $x^{j+1}$ produced by running a Markov chain transition (e.g HMC~\citep{neal2011mcmc}) using samples $x^{j} \sim p_j$ leaves $p_j$ invariant. Computing the importance weights along the sequence of distributions we get,
\begin{equation}
    w_{\text{AIS}}(x^i) = \frac{p_{L-1}(x^i)}{p_L(x^i)}\frac{p_{L-2}(x^i)}{p_{L-1}(x^i)} \dots \frac{\mu_{\text{target}}(x^i)}{p_{1}(x^i)}.
    \label{eqn:ais_weights}
\end{equation}

Plugging the weights $w_{\text{AIS}}$ directly in the previously defined IS estimator gives us the AIS estimator~\citep{neal2001annealed}.

\subsection{Boltzmann generators}
\label{sec:boltzmann_generators}
A Boltzmann generator (BG) samples from $\mu_{\text{target}}$ by combining an exact-likelihood model $q_{\theta}$, typically a Normalizing Flow, and an algorithm to reweight model generated samples using $\mu_{\text{target}}$. During training, a BG's flow is learned by minimizing a convex combination of both the forward and reverse KL-divergence. Thus, to compute the full training loss we need a dataset of ground truth samples from $\mu_{\text{target}}$ as it is required under the forward KL. As a result, BG's are ill-suited in settings with limited or no data as training only with the reverse KL is prone to mode-seeking behavior. Given a trained flow $q_{\theta}$ we can reweight observables $f(x)$ to be under $\mu_{\text{target}}$ by using the IS estimator.

\subsection{Sequential Monte Carlo}
\label{app:smc}

Sequential Monte Carlo (SMC)~\citep{del2006sequential} is an alternative to MCMC methods. It was originally proposed to find approximate solutions to filtering problems (hence, the original name particle filter methods), but was then adopted to sampling problems. In the context of sampling problems, SMC uses a sequence of distributions that map a known distribution to the target. The distributions are constructed in a way that the target distribution is the last distribution in the sequence. One typical choice for the sequence of distributions is the following:

\begin{equation}
    p_j(x) = \mu_{\text{target}}(x)^{\beta_j} p_0(x)^{1-\beta_j},
\end{equation}
where $p_0$ is a known distribution, and $\beta_j$ is a sequence of numbers such that $1 = \beta_0 > \beta_1 > \dots > \beta_n \geq 0$. The SMC algorithm starts by sampling $N$ particles from the initial distribution $p_0(x)$, and then proceeds to iteratively sample new particles from the intermediate distributions $p_j(x)$, for $j \in [1, \dots, n]$, by applying a Markov transition kernel $K_j(x_{t+1} \mid x_t)$ to the particles $x_t \sim p_j(x)$. The particles are then weighted by the importance weights $w_j(x_t) = \frac{p_j(x_t)}{q_j(x_t)}$, and resampled according to the weights. The algorithm terminates when the particles are sampled from the target distribution. SMC can produce high-quality samples even for very complex distributions, particularly when using large numbers of intermediate distributions. Furthermore, the log-partition function can be computed as a by-product of the algorithm, as the average of the log-importance weights.

\subsection{Nested Sampling}

Nested Sampling~\citep{skilling2006nested} is a method that was originally proposed for computing the evidence (or partition function) of a distribution but can also be used for sampling. Like in SMC, in Nested Sampling, we evolve from samples of a known distribution $q(x)$, which we call the prior distribution, to samples of the target distribution $\mu_{\text{target}}(x)$. However, in Nested Sampling we do not use a sequence of distributions, instead, we use a single distribution that is constructed by progressively removing the regions of low-probability mass. The fundamental concept behind nested sampling involves introducing a new variable known as the "cumulative prior mass" or "prior volume," defined as:

\begin{equation}
    \label{eq:prior_volume}
    X(\lambda) = \int_{\mu_{\text{target}}(x) > \lambda} q(x) dx,
\end{equation}
which signifies the proportion of the prior mass with a probability greater than that of the current point. 

The core idea of nested sampling involves the following steps: Initially, a set of $n_{\rm live}$ "live points" is sampled from the prior distribution. The point with the lowest likelihood is identified and moved from the set, becoming a "dead point." Subsequently, it is replaced with a new point drawn from the prior, ensuring its likelihood surpasses that of the removed point. This replacing can be done through various procedures, such as building ellipsoids around the live point~\citep{feroz2013importance}, through slice sampling~\citep{neal2003slice, handley2015polychord}; or through Hamiltonian slice sampling~\citep{speagle2020dynesty,lemos2023improving}. Although exact calculation of $X(\theta)$ for each new point is impractical, it can be approximated by recognizing that, at each iteration, the prior volume contracts by approximately:

\begin{equation}
    \label{eq:delta_x}
    \Delta X \approx \frac{n_{\rm live}}{n_{\rm live} + 1}.
\end{equation}

The algorithm terminates when the prior volume is sufficiently small. The algorithm produces an estimate of the partition function, and samples from the distribution, by re-weighting the killed points by their importance weights:

\begin{equation}
    \label{eq:ns_weights}
    w(x_k) = \frac{\mu_{\text{target}}(x_k) \cdot (X_{k-1} - X_k)}{Z}.
\end{equation}

\section{Related simulation-based and simulation-free diffusion-like samplers}
\label{app:related_work}
In this section, we first describe the basic algorithms behind related work to understand their relative strengths and weaknesses. 

\xhdr{Flow Annealed Bootstrapping~\cite{midgley2022flow} (FAB)} Flow Annealed Bootstrapping uses samples from annealed importance sampling (AIS) to train a normalizing flow model using an $\alpha=2$ divergence. While a continuous normalizing flow~\cite{chen_neural_2018} could be used for more flexibility in architecture, in practice a standard normalizing flow architecture is used, which constrains FAB to invertible architectures in contrast to continuous time models. This is because FAB requires computation of the model likelihood in its loss which requires simulation during training and makes using a continuous normalizing flow a computationally expensive choice. 

We also note that FAB is most effective with a large buffer of AIS samples. Each AIS sampling step takes $\mathcal{O}(M L)$ steps where $M$ is the number of MCMC steps per AIS intermediate distribution and $L$ is the number of AIS levels. Finally, FAB needs to store the importance sampling ratio to compute $w_{\rm AIS}$ which increases its memory footprint. This leads to $L$ importance sampling ratios to supplement the $d$ dimensional data for $\mathcal{O}(L + d)$ memory footprint.

Finally, we note that FAB also finds that a biased objective often leads to better performance in practice than an unbiased objective. In this work, we come to a similar conclusion. That biased estimates can be quite effective in terms of sampling performance as compared to less efficient unbiased estimators.

\xhdr{Simulation-based SDE inference algorithms} The path integral sampler \citep[PIS;][]{zhang2021path}, denoising diffusion sampler \citep[DDS;][]{vargas2023denoising}, and (time-reversed) diffusion sampler \citep[DIS;][]{berner2022optimal} are related algorithms for training a neural SDE \citep{tzen2019neural} to sample from a target distribution. These samplers aim to minimize a variational objective, the KL divergence between two distributions over trajectories: the distribution given by following the denoising SDE starting from the prior and the distribution given by following the fixed noising SDE starting from the target distribution. Such minimization can be achieved with varying choices of time discretizations or integration schemes, but all three methods approximate minimization of the divergence between path space measures in continuous time \citep{richter2023improved} and require numerical integration of the denoising SDE at each iteration of training, giving a computation time linear in the number of steps $L$. This also requires storing the trajectory for gradient computation leading to an $\mathcal{O}(L d)$ memory footprint. 

\xhdr{DEM} The computational complexity of DEM itself is low due to its simulation-free inner loop. With \pdem, sampling from the prior does not require any simulation during training (only during inference)

\xhdr{Sampling Complexity} All methods require simulating forward through the trajectory to generate samples. However, since FAB uses a standard normalizing flow, this is in general cheaper depending on architectural details (but less flexible).

\begin{table}[]
    \caption{Table containing more detailed analysis of method properties during training, and during sampling. $M$ MCMC steps per $L$ annealing steps for FAB, and $d$ dimensionality. These values assume a standard normalizing flow architecture for FAB as used in \citet{midgley2022flow,midgley2023se}.}
    \centering
\resizebox{1\linewidth}{!}{
    \begin{tabular}{lcccccc}
        \toprule
        Method & MCMC Free & Off-Policy & Gradient Time & Proposal Time & Memory & Sampling Time\\
        \midrule
        FAB \citep{midgley2022flow}  &\no  & \yes   & $\mathcal{O}(1)$ & $\mathcal{O}(M L)$ & $\mathcal{O}(L+d)$ & $\mathcal{O}(d)$\\
        PIS \citep{zhang2021path} &\yes & \no       & $\mathcal{O}(L)$ & $\mathcal{O}(L)$ & $\mathcal{O}(L d)$ & $\mathcal{O}(L d)$\\
        DDS \citep{vargas2023denoising} &\yes & \no & $\mathcal{O}(L)$ & $\mathcal{O}(L)$ & $\mathcal{O}(L d)$& $\mathcal{O}(L d)$\\
        DIS \citep{berner2022optimal}  &\yes & \no & $\mathcal{O}(L)$& $\mathcal{O}(L)$ & $\mathcal{O}(L d)$& $\mathcal{O}(L d)$\\
        \pdem (ours) & \yes & \yes & $\mathcal{O}(1)$& $\mathcal{O}(1)$ & $\mathcal{O}(d)$ & $\mathcal{O}(L d)$\\
        \name (ours) &\yes & \maybe & $\mathcal{O}(1)$ & $\mathcal{O}(L)$ & $\mathcal{O}(d)$& $\mathcal{O}(L d)$\\
        \bottomrule
    \end{tabular}
}
    \label{tab:complexity_full}
\end{table}

\section{Additional Details on the Experiments}
\label{app:experimental_setup}

\subsection{Experimental Setup}
\label{app:experimental_setup:architecture}
In this section, we detail the exact setup for all of our experiments. For each experimental task and method (besides FAB for GMM, DW-4, and LJ-13 for which we use the hyperparameters reported best in \citet{midgley2022flow}, \citet{midgley2023se} respectively) we performed a grid search to find the best hyperparameters and evaluated each setting over three random seeds.   For \name we use a geometric noise schedule $\sigma(t) = \sigma_{min}(\frac{\sigma_{max}}{\sigma_{min}})^t$ and tune over learning rate as well as $\sigma_{min}$ and $\sigma_{max}$. For PIS we tune over the learning rate and the coefficient of the Brownian motion. For DDS we tune over using their proposed exponential or Euler integration, $\sigma_{max}$ and $\alpha_{max}$ when using the exponential integration, and $\beta_{min}$ and $\beta_{max}$ for Euler integration. For FAB on LJ-55 we tune over learning rate and the number of intermediate distributions. All networks were optimized using Adam and were performed on NVIDIA A100 GPUs with 40GB of VRAM. We commit to releasing our code upon publication.

We provide further details of our setup for each of the experiments below:

\xhdr{GMM} All models use an MLP with sinusoidal and positional embeddings. The MLP has 3 layers of size 128 as well as positional embeddings of size 128. Both \name and FAB use a replay buffer of max length 10000 which is prioritized by energy for FAB and unprioritized for \name. All methods were trained with Adam. 

For training \name, the generated data was in the range [-1, 1] so to calculate the energy it was scaled appropriately by unnormalizing by a factor of 50. \name was trained with a geometric noise schedule with $\sigma_{\rm min} = 1e-5, \sigma_{\rm max} = 1$, $K=500$ samples for computing the regression target $\mathcal{S}_K$ and we clipped the norm of $\mathcal{S}_K$ to 70. \name was trained with a learning rate of $5e-4$.  All other methods did not normalize the data. PIS was trained with a learning rate of $5e-4$ and a Brownian motion coefficient of $30$. DDS was trained with a learning rate of $5e-4$ and used the exponential integrator proposed in \citet{vargas2023denoising}. We use $\alpha = 0.3$ and $\sigma = 30$ for the exponential integration. FAB was trained following exactly the settings used in \citet{midgley2022flow}.

\xhdr{DW-4}
All models besides FAB used an EGNN with 3 message-passing layers and a 2-hidden layer MLP of size 128. For FAB, we used the $\sethree$-augmented coupling flow architecture from \citet{midgley2023se} due to its requirement of an invertible architecture. \name was trained with a geometric noise schedule with $\sigma_{min}=1e-5$ and $\sigma_{max}=3$, a learning rate of $1e-3$, and $K=1000$ samples for computing the regression target $\mathcal{S}_K$ and we clipped the regression target to a max norm of $20$. PIS was trained with a learning rate of $5e-4$ and we used $1$ for the coefficient of the Brownian motion. DDS was trained with a learning rate of $5e-3$ and we used Euler integration with $\beta_{min}=0.5,\beta_{max}=1.5$. FAB was trained following exactly the settings used in \citet{midgley2023se}.

\xhdr{LJ-13}
All models besides FAB used an EGNN with 5 hidden layers and hidden layer size 128 while FAB used the architecture from \citet{klein2023equivariant}. \name was trained with a geometric noise schedule with $\sigma_{min}=0.01$ and $\sigma_{max}=2$, a learning rate of $1e-3$, $K=1000$ samples for the regression target $\mathcal{S}_K$ and clipped the regression target to a max norm of $20$. PIS was trained with a learning rate of $1e-4$ and a Brownian motion coefficient of 1. DDS was trained with a learning rate of $5e-3$ and Euler integration with $\beta_{max}=0.5$ and $\beta_{min}=0.01$. FAB was trained following exactly the settings used in \citet{midgley2023se} using their $\sethree$-augmented coupling flow architecture with spherical projection.

\xhdr{LJ-55}
All models besides FAB used an EGNN with 5 hidden layers and hidden layer size 128 while FAB used the architecture from \citet{klein2023equivariant}. \name was trained with a geometric noise schedule with $\sigma_{min}=0.5$ and $\sigma_{max}=4$, a learning rate of $1e-3$, $K=100$ samples for the regression target $\mathcal{S}_K$ and clipped the regression target to a max norm of $20$. PIS was trained with a learning rate of $1e-4$ and a Brownian motion coefficient of 1. DDS was trained with a learning rate of $5e-3$ and Euler integration with $\beta_{max}=0.5$ and $\beta_{min}=0.01$. FAB was trained with the $\sethree$-augmented coupling flow architecture from \citet{midgley2023se} with spherical projection, 16 intermediate distributions, a learning rate of $2e-5$, and a batch size of $8$ (the most that fit in GPU memory of the 40GB NVIDIA A100 GPUs used).

\subsection{Metrics reported in Table~\ref{tab:results_wide} and Table~\ref{tab:additional_metrics}}\label{app:experimental_setup:results_wide}
\autoref{tab:results_wide} depicts the main experimental sample quality results for various methods on all datasets and additional metrics are reported in \autoref{tab:additional_metrics}. In this section, we explain the methodology for each experiment. We note that it was difficult to train PIS and DDS for the high-dimensional Lennard-Jones tasks. 2/3 seeds of DDS diverged for LJ-13.

\xhdr{Negative Log Likelihood (NLL)} The negative log-likelihood measures how likely a test dataset is under a model. For different model types, there are different methods to numerically calculate or approximate the NLL of a test sample. In this work, we use the exact likelihood from a continuous normalizing flow (CNF) model. We train this CNF using optimal transport flow matching on samples from each model (OT-CFM)~\citep{tong_conditional_2023}. Then this CNF can subsequently be used to calculate the likelihood of a test set. This allows us to compare the sample quality of various model architectures fairly using the same model architecture, training regime, and numerical likelihood approximation independent of the sampler form. 

For the CNF with flow function $f: ([0,1], \R^d) \to \R^d$, we use the exact estimator of the likelihood, \ie, for a test sample $x_0$, its likelihood $p_{\rm model}(x)$ can be estimated as 
\begin{equation}
    \log p_{\rm model}(x) = \log p_{\rm prior}(x_0) + \int_{1}^0 - \text{Tr} \left ( \frac{d f}{d x_t} \right ) dt \nonumber
\end{equation}
where $x(t) = x_1 - \int_{1}^0 f(t, x) dt$ using the continuous-time change of variables formula. We note that it is important to choose the integration method correctly, as the model is only invertible in continuous time. If too few steps are taken then we observe inaccurate NLL values. Therefore we use the 5th order Dormand-Prince (dopri5) adaptive step size solver with tolerances \texttt{atol=rtol=}$10^{-3}$ for GMM and \texttt{atol=rtol=}$10^{-5}$ for DW-4, LJ-13, and LJ-55. This keeps the number of function evaluations per integration around 100 in practice, but results in a much more accurate and repeatable NLL value.

For more complicated datasets such as LJ13 and LJ55, we find that training a CNF on the output of some samples degrades the NLL performance of the CNF. We refer to a sampler as \textit{diverged} if the negative log-likelihood of the CNF trained on its output is worse (higher) than an untrained sampler. For LJ13 and LJ55 respectively, an untrained CNF achieves a negative log likelihood of 60.32 and 230.53. In our reported results with mean and standard deviation, we exclude these values from the aggregation. We exclude values for LJ55 for the PIS, DDS, and pDEM models where training the CFM on samples from the sampling models do not improve the NLL of the CFM over random initialization. 

There are other methods to compute or approximate NLL. In \autoref{tab:nll_comparison} we show that CFM is a relatively good estimator of the NLL as compared to the native estimations used in FAB and PIS respectively. This supports our use of a standardized model and training procedure for NLL computation.

In practice, we train CFM on 100k samples on GMM, DW-4, LJ-13 tasks and 10k samples on the LJ-55 task for all models. We train on fewer samples from LJ-55 due to the cost of sampling in this high-dimensional setting. It may be possible to achieve better NLL values with more samples (particularly in LJ-55) or with better architectures, due to our standardized pipeline, this does not affect the comparison between samplers evaluated in this work.

\xhdr{2-Wasserstein distance $\mathcal{W}_2$} We also use the standard 2-Wasserstein distance between empirical samples from the sampler and the ground truth dataset. The 2-Wasserstein distance is defined as 
\begin{equation}
    \mathcal{W}_2(\mu, \nu) = \left ( \inf_\pi \int \pi(x,y) d(x,y)^2 dx dy \right )^{\frac{1}{2}}
\end{equation}
where $\pi$ is the transport plan with marginals constrained to $\mu$ and $\nu$ respectively. In practice, we use the Hungarian algorithm as implemented in the Python optimal transport package (POT)~\citep{flamary2021pot} to solve this optimization for discrete samples. For simplicity, we use the Euclidean ground distance $d(x,y)=\| x - y\|_2$.

\xhdr{Effective Sample Size (ESS)} To measure ESS we first evaluate log importance weights given by using our model $p_{model}(x)$ as a proposal.  We estimate the ESS by

\begin{equation*}
    ESS = \frac{n}{\sum_{i=1}^n w_i^2}
\end{equation*}

where $w_i = \frac{\exp(-\mathcal{E}(x_i)) / p_{model}(x_i)}{\sum_{j=1}^n \exp(-\mathcal{E}(x_j)) / p_{model}(x_j)}$, which is equivalent to the Softmax of the log-probability ratios. We note that this is sometimes known as the normalized ESS, as we normalize the effective sample size to the fraction of the test set size. 

\xhdr{Total Variation} The total variation metric on the Gaussian mixture model dataset is taken over 200 bins in each dimension (200$^2$) total bins. This is possible in low dimensions but does not scale well to high dimensions, requiring an exponential number of bins. Therefore for the larger equivariant datasets, we take the total variation distance over the distribution of the interatomic distances of the particles, see \eg, \autoref{fig:main_energy_figs}, which scales well with dimension and shows how well the distribution of energies matches that of the test data. 

\xhdr{Log partition function ( $\log{Z}$ )}
To compute $\log Z$ we use an importance sampling estimate with the proposal density, $q(x)$, given by the OT-CFM model as
\begin{align}
    \log Z &= \log \mathbb{E}_{x \sim q(x)}\left[\frac{\exp(-\mathcal{E}(x))}{q(x)}\right] \\
    &\geq \mathbb{E}_{x \sim q(x)}[-\mathcal{E}(x) - \log q(x)]
\end{align}
which yields a lower bound on the true log partition function. Note that for GMM we know the true value of $\log{Z}$, however, this isn't the case for the equivariant tasks. For those tasks, to compare the values of $\log{Z}$ from the different methods, we use the fact that our estimate is a lower bound of the true value and favour the method with the largest estimate.  

\subsection{Timing experiment setup}
To compute the training times in Table \ref{tab:train_time} we first measure the time per iteration while excluding all computations used for evaluation. Next, we determined the number of training steps required for each method to converge based on training loss. Finally, we reported the number of iterations taken to converge multiplied by the number of iterations per second for each method in Table \ref{tab:train_time}.

\subsection{Task details}
\label{app:experimental_setup:target_densities}

\subsubsection{Gaussian Mixture Model}
We use a 40 Gaussian mixture density in 2 dimensions as proposed by \citet{midgley2022flow}. This density consists of a mixture of 40 evenly weighted Gaussians with identical covariances \begin{equation}
    \Sigma = \begin{pmatrix}
40 & 0\\ 
0 & 40
\end{pmatrix}
\end{equation}
and $\mu_i$ are uniformly distributed over the $-40$ to $40$ box, \ie, $\mu_i \sim \mathcal{U}(-40, 40)^2$.
\begin{equation}
    p_{\rm gmm}(x) = \frac{1}{40} \sum_{i=1}^{40} \mathcal{N}(x ; \mu_i, \Sigma)
\end{equation}

We use a test set of size 1000 sampled with \textsc{torch.random.seed(0)} following prior work.

\subsubsection{DW-4}
\begin{figure}[t]
    \captionsetup[subfigure]{aboveskip=-1pt,belowskip=-1pt}
    \begin{subfigure}{0.49\linewidth}
        \centering
        \includegraphics[width=\linewidth]{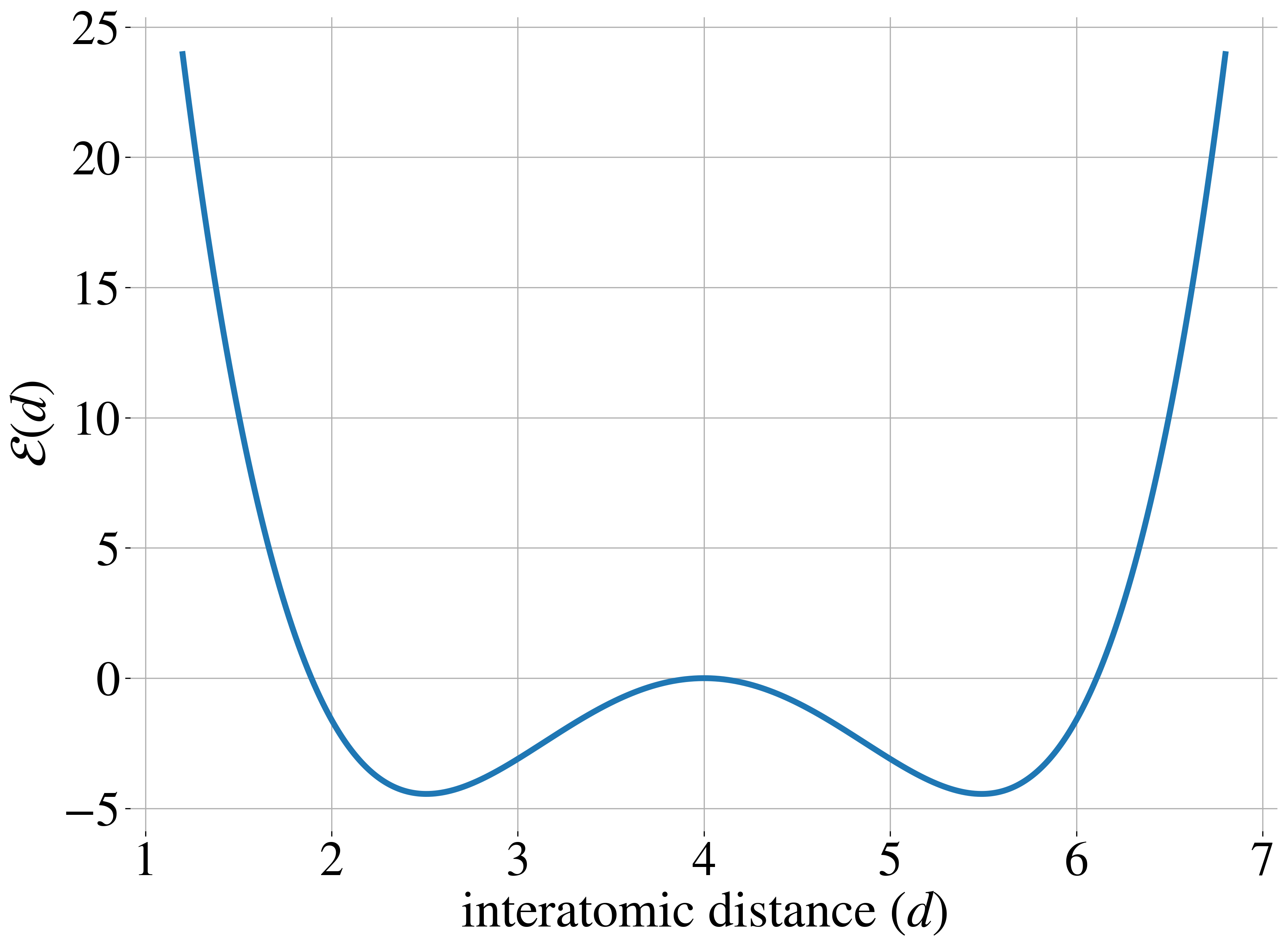}
        \label{fig:energy_vs_dist:a}
    \end{subfigure}
    \begin{subfigure}{.49\linewidth}
    \centering
        \includegraphics[width=\linewidth]{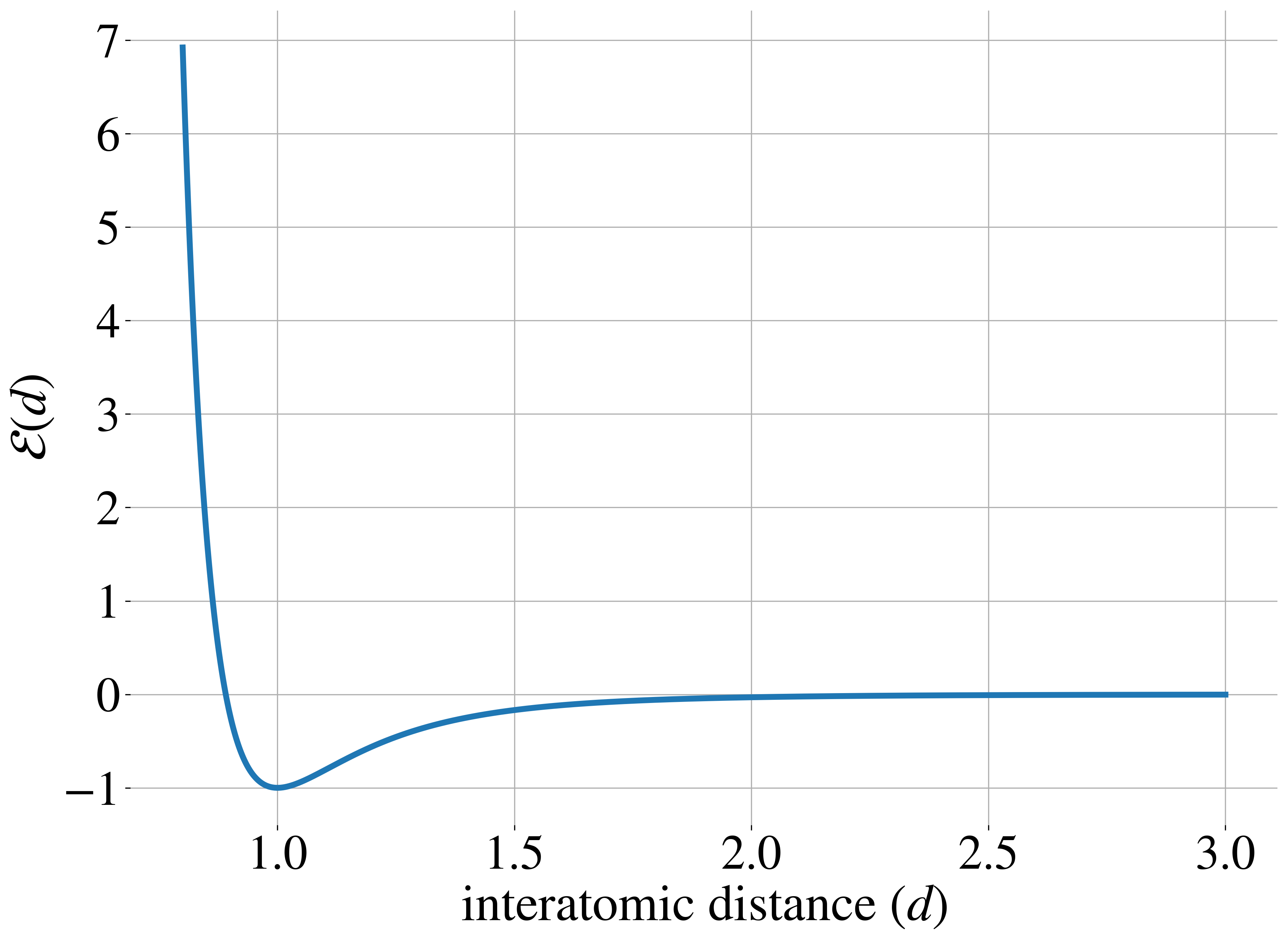}
        \label{fig:energy_vs_dist:b}
    \end{subfigure}
    \vspace{-10pt}
    \caption{\small
    Energies of (left) double-well potential and (right) Lennard-Jones potential ($\mathcal{E}^{LJ}$) as a function of the interatomic distance between the particles. 
    }
    \vspace{-10pt}
    \label{fig:energy_vs_dist}
\end{figure}

The energy function for the DW-4 dataset was introduced in \citet{kohler2020equivariant} and corresponds to a system of 4 particles in a 2-dimensional space. The system is governed by a double-well potential based on the pairwise distances of the particles. For a system of 4 particles, $x = \{x_1, \dots, x_4\}$, the energy is depicted in \cref{fig:energy_vs_dist:a} and is given by:
\begin{equation}
    \gE^{\rm DW}(x) = \frac{1}{2 \tau} \sum_{ij} a (d_{ij} - d_0) + b (d_{ij} - d_0)^2 + c (d_{ij} - d_0)^4
\end{equation}
where $d_{ij} = \|x_i - x_j\|_2$ is the Euclidean distance between particles $i$ and $j$. Following previous work, we set $a=0$, $b=-4$, $c=0.9$ and the temperature parameter $\tau=1$. To evaluate the efficacy of our samples we use a validation and test set from the the MCMC samples in \citet{klein2023equivariant} as the ``Ground truth'' samples. We note that this is not necessarily a perfect ground truth, but we believe it is reasonable to use for our purposes. Previous work has evaluated performance based on a dataset from \citet{garcia2021n}. However, we note that this dataset is biased as its test set partition is generated from a single MCMC chain. As such, we omit this dataset and evaluate only on the data from \citet{klein2023equivariant}.

\subsubsection{Lennard-Jones Potential}
\label{ref:app_lj}
The Lennard-Jones (LJ) potential is an intermolecular potential which models repulsive and attractive interactions of non-bonding atoms or molecules. The energy is based on the distance of interacting particles and is given by:
\begin{equation}\label{eq:lj}
    \gE^{\rm LJ}(x) = \frac{\eps}{2 \tau} \sum_{ij} \left( \left( \frac{r_m}{d_{ij}}\right)^6 - \left( \frac{r_m}{d_{ij}}\right)^{12} \right)
\end{equation}
where $d_{ij}=\|x_i - x_j\|_2$ is the Euclidean distance between particles $i$ and $j$, and $r_m$, $\tau$, $\epsilon$ and $c$ are physical constants. As in \citet{kohler2020equivariant}, we also use a harmonic potential:
\begin{equation}
    \gE^{\rm osc}(x) = \frac{1}{2} \sum_{i}||x_i - x_{\rm COM}||^2
\end{equation}
where $x_{\rm COM}$ refers to the center of mass of the system. Therefore, the final energy is then  $\gE^{Tot} = \gE^{\rm LJ}(x) + c \gE^{\rm osc}(x)$, for $c$ the oscillator scale. As in previous work, we use $r_m=1$, $\tau=1$, $\eps=1$ and $c = 0.5$. We note that this task is difficult not only because of its dimensionality but also because of the high magnitude of its score. As depicted in \autoref{fig:energy_vs_dist} (left) and apparent from \eqref{eq:lj}, the score explodes as any $d_{ij} \to 0$, making this task particularly challenging in higher dimensions.

LJ-13 refers to the system of 13 particles, $x = \{ x_1, \dots, x_{13} \}$ with 3 dimensions each, resulting in a task with dimensionality $d = 39$. LJ-55 meanwhile refers to a system of 55 particles, $x=\{x_1, \dots, x_{55} \}$ with 3 dimensions each, resulting in a high dimensional task with dimensionality $d = 165$. For the experimental results, we evaluate using the MCMC samples from \citet{klein2023equivariant}. As with DW-4, previous work evaluated performance based on a dataset from \citet{garcia2021n}. However, this dataset is also biased with the test set partition generated from a single MCMC chain and is generated with $\mathcal{E}^{LJ} / 2$ as the sum is only calculated for $i < j$. Therefore, we only evaluate the models on the data from \citet{klein2023equivariant}.

\section{Additional results}
\label{app:additional_results}

\subsection{Additional metrics}
\label{app:additional_metrics}
We report here additional quantitative metrics to supplement the main results presented in \autoref{sec:main_results}.  In \autoref{tab:additional_metrics}, we report the Total Variation (TV) as well as the log partition function ($\log{Z}$) values.

\begin{table}[h]
\caption{\small Sampler performance with mean $\pm$ standard deviation over 3 seeds for Total Variation (TV), and log partition function ($\log Z$). $*$ indicates divergent training and entries with $^\dagger$ refer to settings where only 1 of 3 runs converged. For TV, we \textbf{bold} via Welch's two-sample t-test $p < 0.1$. For $\log{Z}$, we \textbf{bold} the method with the largest value, as our estimate is a lower bound on the true $\log{Z}$.}
\label{tab:additional_metrics}
\resizebox{1\linewidth}{!}{
\begin{tabular}{@{}lcccccccc}
    \toprule
    Energy $\rightarrow$ & \multicolumn2c{GMM ($d=2$)} & \multicolumn2c{DW-4 ($d=8$)} & \multicolumn2c{LJ-13 ($d=39$)} & \multicolumn2c{LJ-55 ($d=165$)} \\
    \cmidrule(lr){2-3}\cmidrule(lr){4-5}\cmidrule(lr){6-7}\cmidrule(lr){8-9}
    Algorithm $\downarrow$ & TV & $\log{Z}$ & TV & $\log{Z}$ & TV & $\log{Z}$ & TV & $\log{Z}$ \\
    \midrule
    FAB~\citep{midgley2022flow} 
    &0.88\std{0.02} & -1.165 \std{0.164}
    &\textbf{0.09\std{0.00}} & 29.602 \std{0.019}
    &\textbf{0.04\std{0.00}} & 4.35\std{0.01} 
    &0.24\std{0.09} & 32.809$^\dagger$ \\
    PIS~\citep{zhang2021path}
    & 0.92\std{0.01} & -2.243 \std{0.070}
    & \textbf{0.09\std{0.00}} & 29.599 \std{0.009}
    & 0.25\std{0.01} & 46.685 \std{1.471}
    & $*$            & $*$ \\
    DDS~\citep{vargas2023denoising}
    & \textbf{0.82\std{0.02}} & -0.358 \std{0.209}
    & 0.16\std{0.01} & 28.382 \std{0.158}
    & $*$            & $*$ 
    & $*$            & $*$  \\
    \pdem (ours)
    & \textbf{0.82\std{0.02}} &  -0.370 \std{0.005}
    & 0.13\std{0.00} & 29.191 \std{0.036}
    & 0.06\std{0.02} & 32.450 \std 3.191 
    & $*$            & $*$ \\
    \name (ours) 
    & \textbf{0.82\std{0.01}} & \textbf{-0.340 \std{0.075}}
    & \textbf{0.10\std{0.01}} & \textbf{29.567 \std{0.014}} 
    & \textbf{0.04\std{0.01}} & \textbf{49.969 \std{2.784}}
    & \textbf{0.09\std{0.01}} & \textbf{273.167 \std{22.226}}\\
    \bottomrule
    \end{tabular}
    }
\end{table}

In \autoref{tab:nll_comparison} we report the NLL values that are computed between the original method and the CFM model that is trained using the samples of each method.

\begin{table}[ht]
    \centering
    \caption{This table compares native vs.\ negative log-likelihood estimation by retraining a flow matching model on samples. FAB-Native directly admits a likelihood calculation through its invertible architecture. PIS-Native uses a stochastic estimation of an upper bound on the NLL. 231 is the negative log-likelihood of a randomly initialized CFM model, therefore any value larger than this is worse than random. We report NLL values worse than the initialization for CFM as $\ge$ 231.}
    \begin{tabular}{@{}l r r r r}
    \toprule
    Algorithm $\downarrow$ Energy $\rightarrow$ & GMM & DW-4 & LJ-13 & LJ-55\\
    \midrule
    FAB-Native     & 7.12\std{0.12}  & 7.19\std{0.06}   & 17.40\std{0.10}  & 1355\std{1379} \\ 
    FAB-CFM        & 7.14\std{0.01}  & 7.16\std{0.01}   & 17.52\std{0.17}  & 200.32\std{62.30} \\ 
    \midrule
    PIS-Native                             & 7.92\std{0.06}  & 7.31\std{0.02}   & 47.783\std{2.283} & 449.794\std{476.535} \\
    PIS-CFM              & 7.72\std{0.03} & 7.19\std{0.01}  & 47.05\std{12.46} & $\ge231$\\
    \bottomrule
    \end{tabular}
    \label{tab:nll_comparison}
\end{table}

\begin{figure}[ht]
    \centering
    \includegraphics[width=0.48\linewidth]{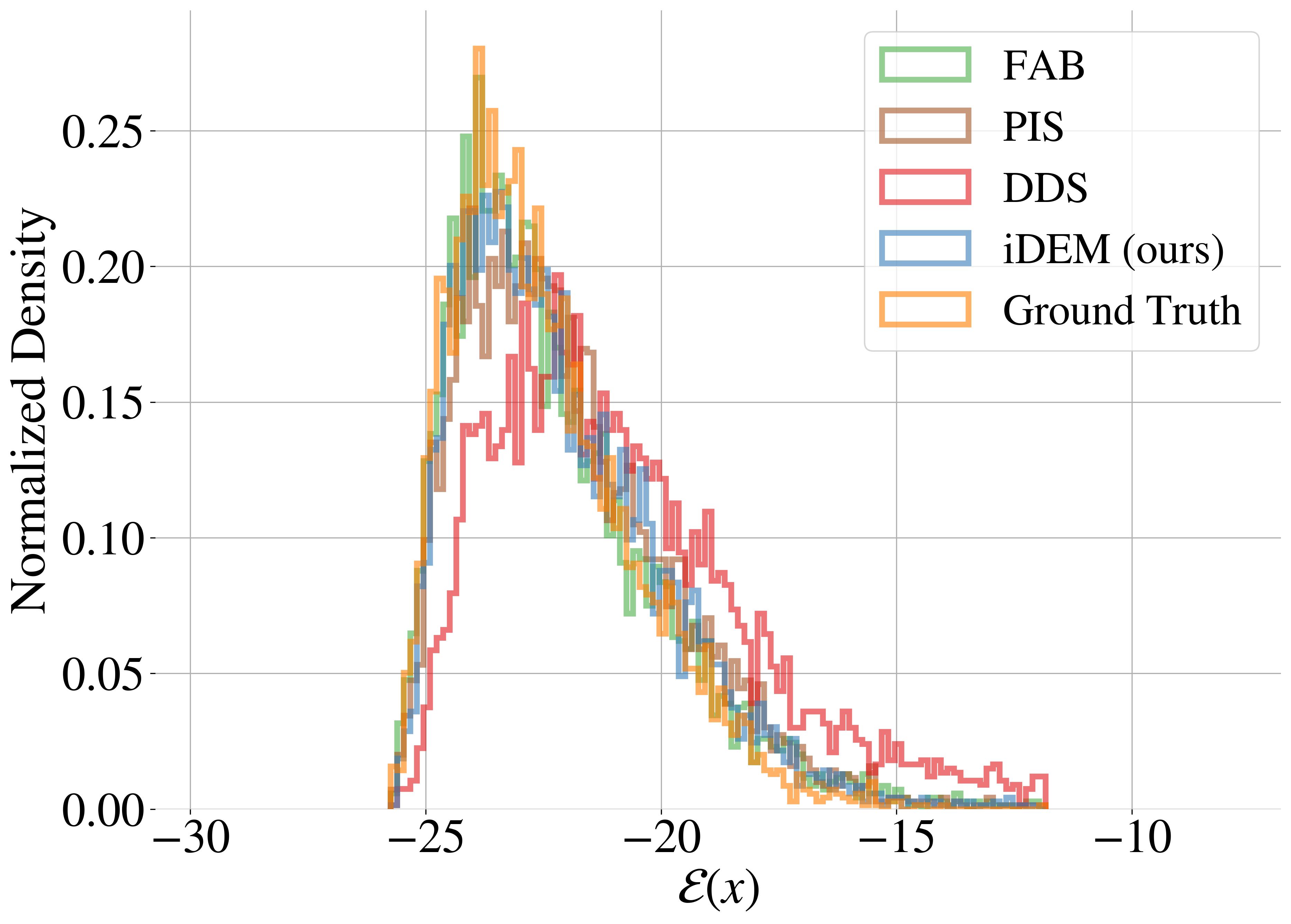}
    \caption{ \small
        Comparison of the ground truth energy histograms of DW-4 in relation to energies of samples generated from \name and baseline methods.
    }
    \label{fig:dw4_energy_fig}
\end{figure}

In \autoref{fig:dw4_energy_fig} we plot the energy histogram of the DW-4 system in comparison to all the methods. We observe that \name and FAB achieve similar performance and match the ground truth energy. PIS and DDM also are able to learn using this energy but are noticeably worse than \name and FAB.

\subsection{Interatomic distances}
\label{app:interatomic_distance}

\begin{figure}[h]
    \centering
    \includegraphics[width=0.33\linewidth]{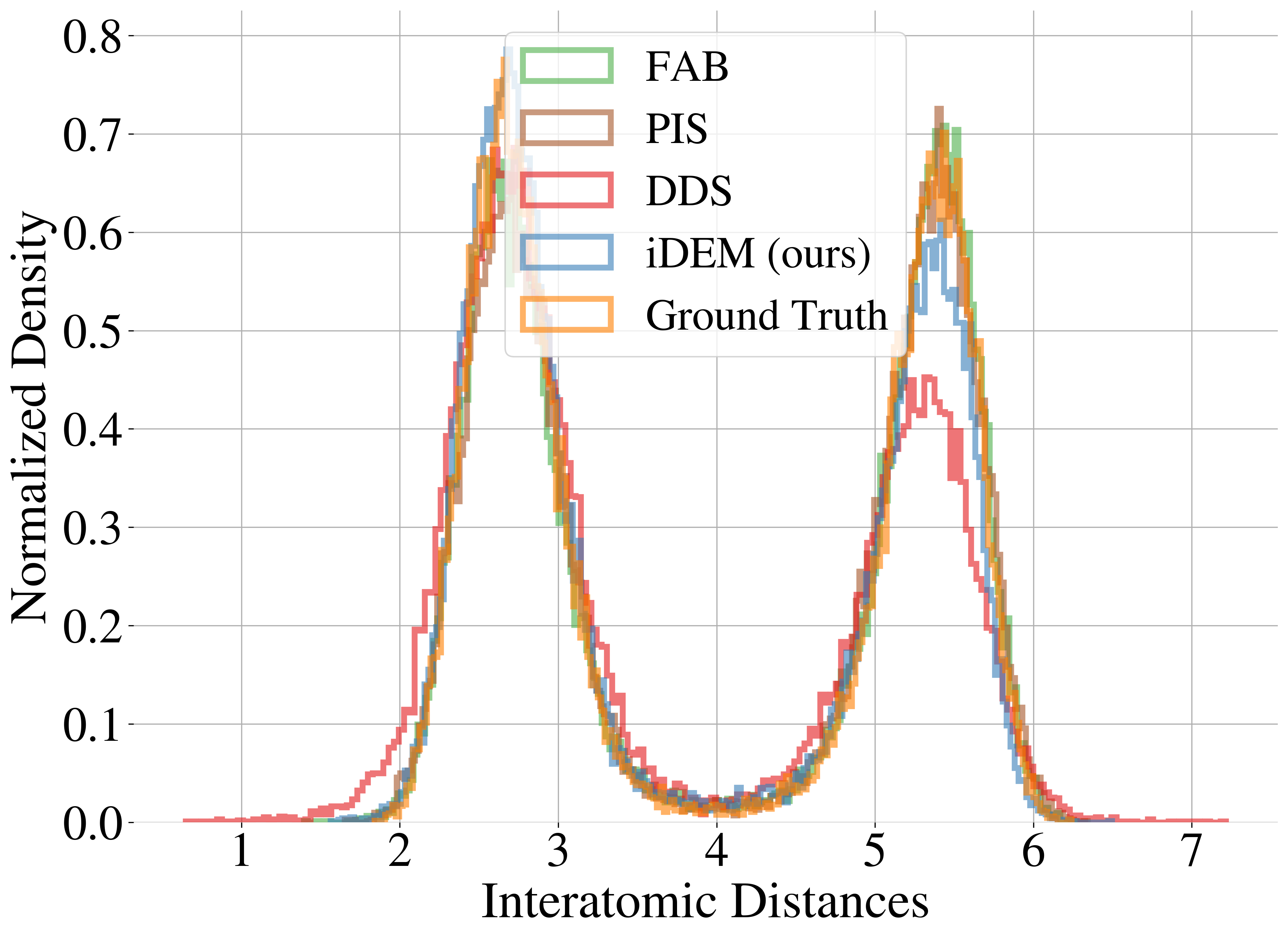}
    \includegraphics[width=0.33\linewidth]{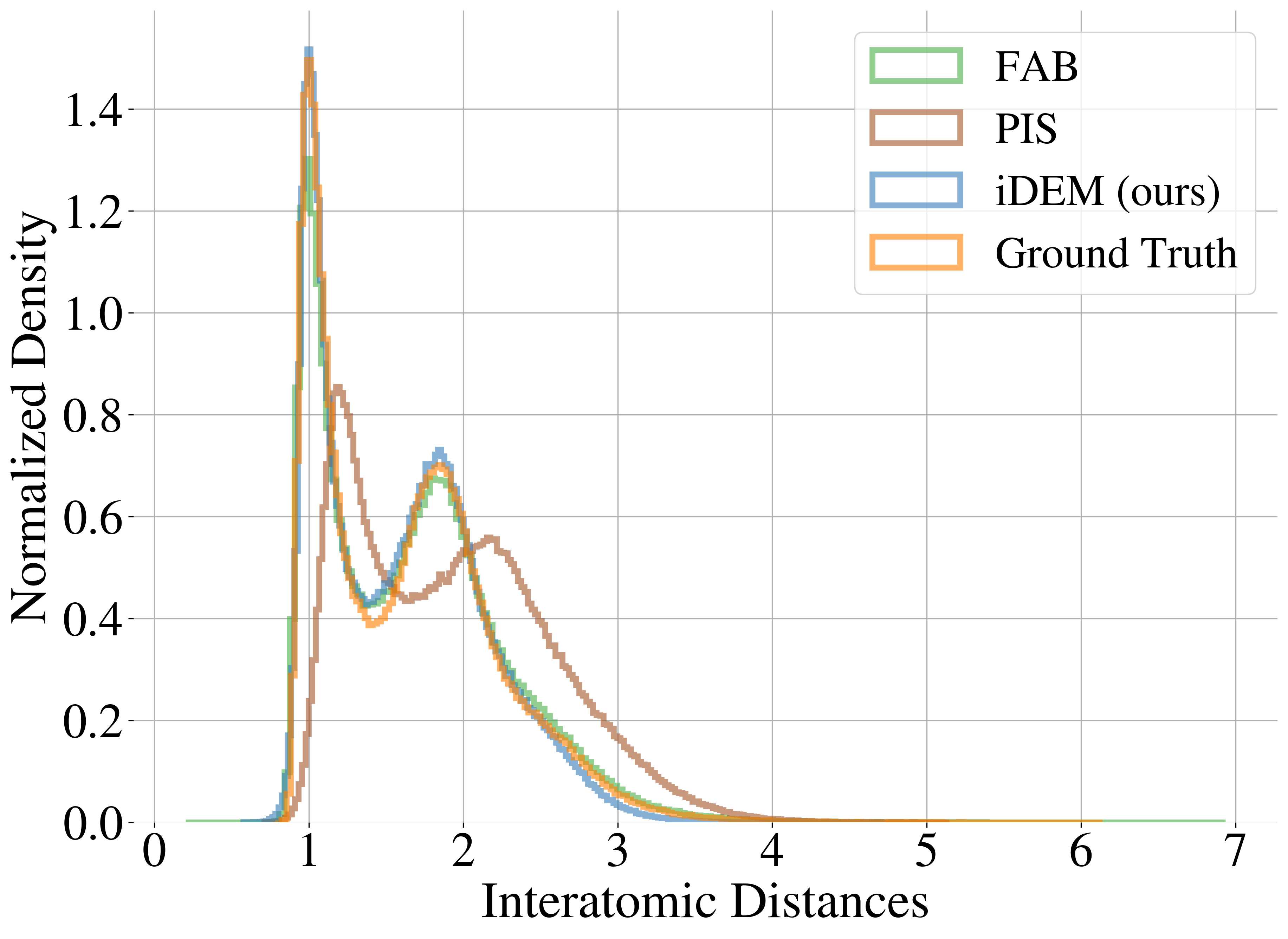}
    \includegraphics[width=0.33\linewidth]{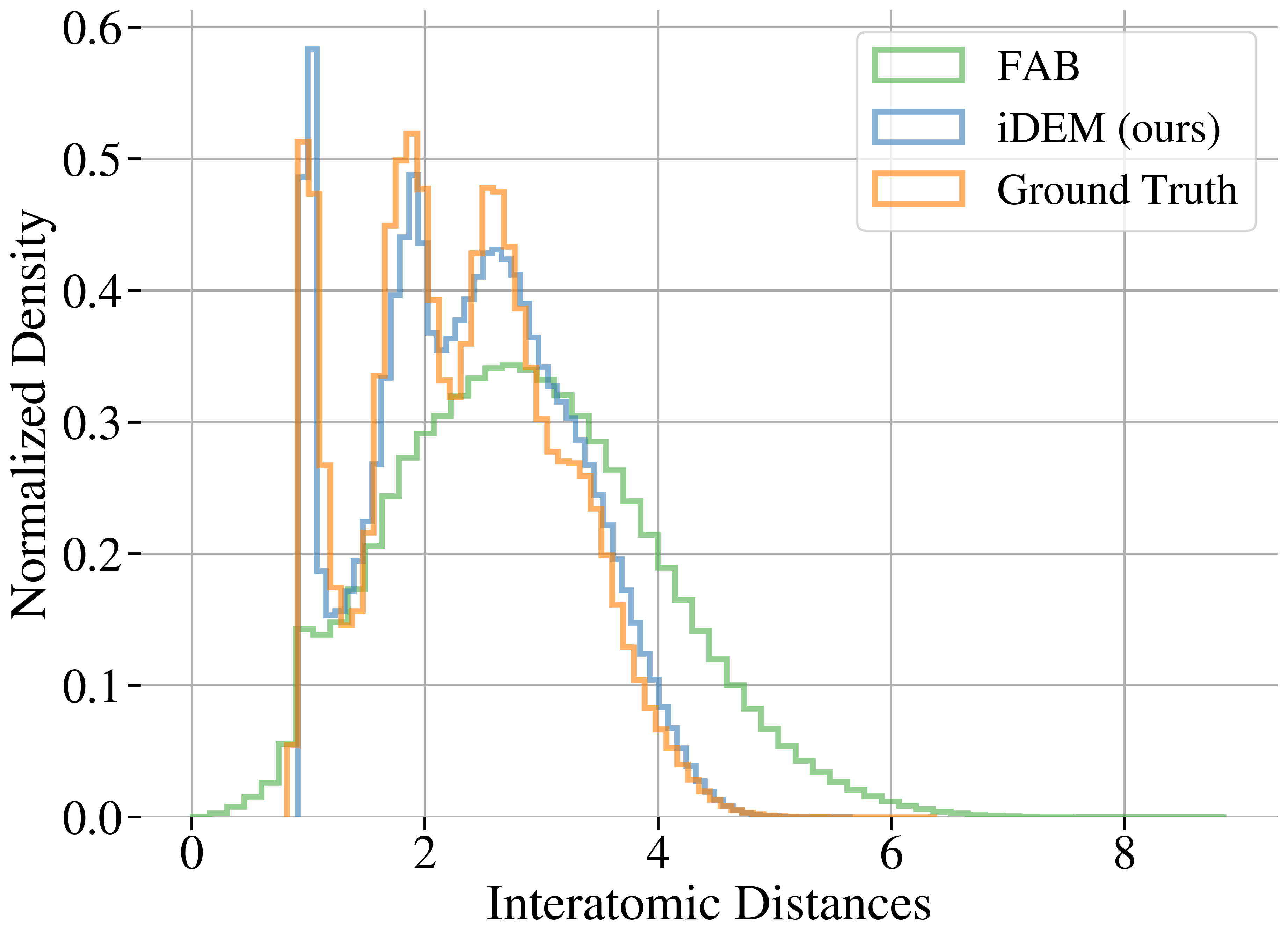}
    \caption{
        Interatomic distances for DW-4 (\textbf{Left}), LJ-13 (\textbf{mid}), and LJ-55 (\textbf{right}) of the ground truth versus the model-generated samples. Note that PIS is omitted from the LJ-55 plot and DDS is omitted from both LJ-13 and LJ-55 plots as the samples of these models diverged in these settings.
    }
    \label{fig:interatomic_dist}
\end{figure}

We report the interatomic distances of the ground truth system and model-generated samples for the DW-4, LJ-13, and LJ-55 tasks in \autoref{fig:interatomic_dist}. We find the highest agreement with the ground truth method and \name with the differences between baselines increasing with the complexity of the dataset.

\subsection{Further ablations}
\label{app:further_ablations}

\xhdr{Different choices of $\gS_K$} While in practice we use the biased estimate (\ref{eq:mc_score_esimate}) one could consider using other estimates of $\nabla_{x_t} \log p_t(x_t)$ as a regression target. For example, instead of applying the LogSumExp trick as is done in (\ref{eq:mc_score_esimate}) one could work directly with the ratio  estimate (\ref{eq:score_as_expectation_boltzmann}), namely
\begin{align*}
    \nabla_{x_t} \log p_t(x_t) &= \frac{\mathbb{E}_{x_{0|t}\sim \mathcal{N}(x_t, \sigma_t^2)}[\nabla \exp(-\mathcal{E}(x_{0|t}))]}{\mathbb{E}_{x_{0|t}\sim \mathcal{N}(x_t, \sigma_t^2)}[\exp(-\mathcal{E}(x_{0|t}))]} \\
    &\approx \frac{\frac1K\sum_i\nabla\exp(-\gE(x_{0\mid t}^{(i)}))}{\frac1K\sum_i\exp(-\gE(x_{0\mid t}^{(i)}))}\nonumber\\
    x_{0\mid t}^{(1)},\dots,x_{0 \mid t}^{(K)} & \sim\gN(x_t,\sigma^2_t),\nonumber
\end{align*}

Inherent in this estimate are drawbacks, chief among them that we must work with the non-log scale energies $\exp(-\mathcal{E}(x_{0|t}))$ directly which can often be either extremely large or extremely small depending on the landscape of $\mathcal{E}$. This can cause large variance estimates and numerical instability if one is unlucky.

Another possible estimate for $\nabla \log p_t(x_t)$ is inspired by Jensen's inequality. In particular, if we write $\nabla \log p_t(x_t) = \nabla \log E_{x_{0|t} \sim \mathcal{N}(x_t,\sigma_t^2)}[\exp(-\mathcal{E}(x_{0|t})]$ we can push the $\log$ inside the expectation and observe that

\begin{align*}
    \nabla_{x_t} \log p_t(x_t) &\approx \nabla_{x_t} \mathbb{E}_{x_{0|t}\sim \mathcal{N}(x_t,\sigma_t^2)}[-\mathcal{E}(x_{0|t})] \\
    &= \nabla_{x_t} \int -\mathcal{N}(x_{0|t}; x_t, \sigma_t^2) \mathcal{E}(x_{0|t})dx_{0|t} \\
    &= -\int \nabla_{x_t} \mathcal{N}(x_{0|t}; x_t, \sigma_t^2) \mathcal{E}(x_{0|t})dx_{0|t} \\
    &= -\int \mathcal{N}(x_{0|t}; x_t, \sigma_t^2) \mathcal{E}(x_{0|t}) \nabla_{x_t} \log \mathcal{N}(x_{0|t}; x_t, \sigma_t^2) dx_{0|t} \\
    &= -\frac{1}{\sigma_t^2}\mathbb{E}_{x_{0|t}\sim \mathcal{N}(x_t, \sigma_t^2)}\left[\E(x_{0|t})(x_{0|t}-x_t)\right],
\end{align*}

This estimate is reminiscent of the estimate used in typical score matching where we have access to samples from $p_0$. Moreover, the estimate does not require access to the gradient of $\E$ making it an attractive option if it is indeed faithful to the true score $\nabla_{x_t} \log p_t(x_t)$.

\begin{figure}[h]
    \centering
    \includegraphics[width=0.48\linewidth]{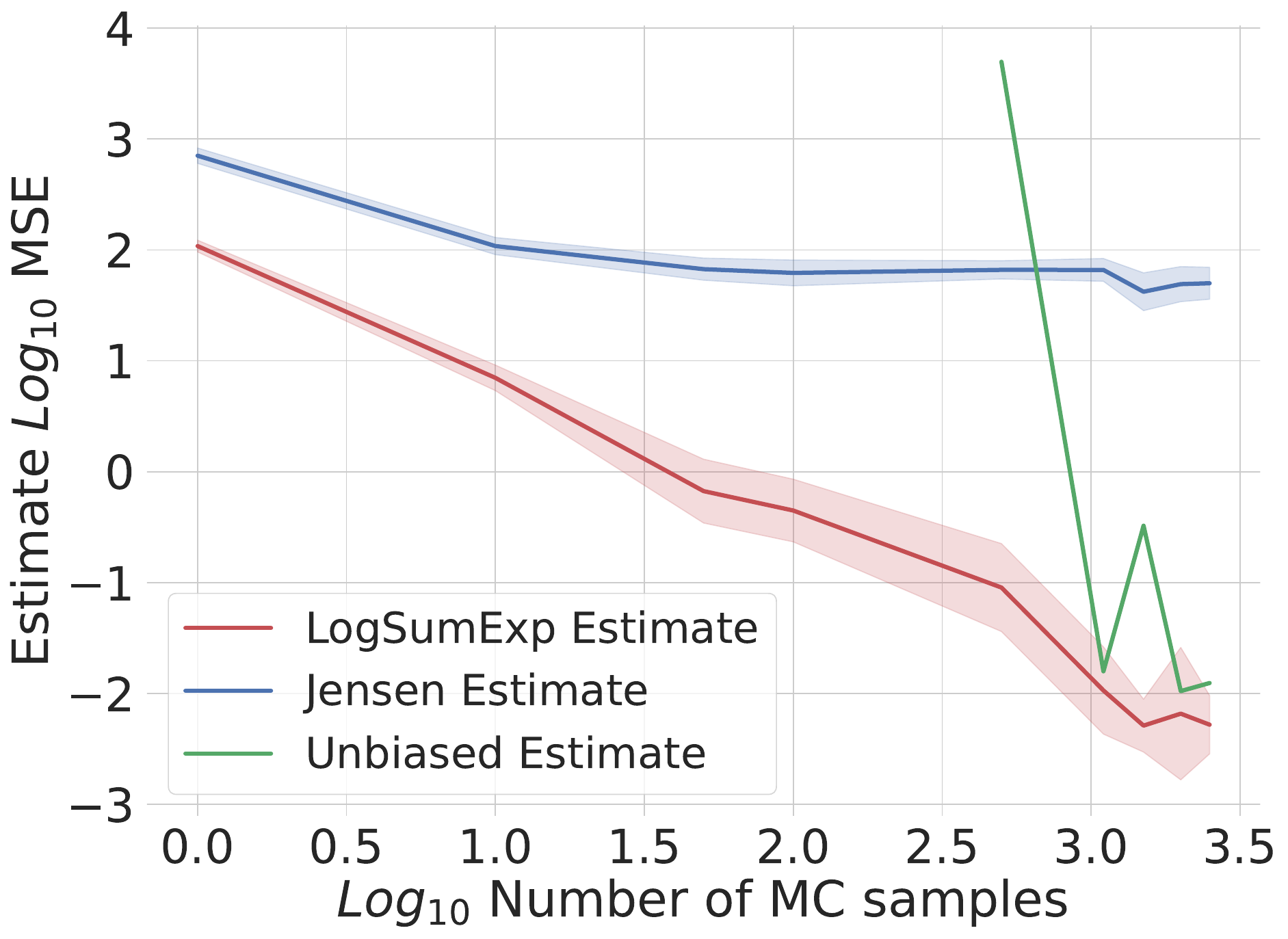}
    \vspace{-10pt}
    \caption{
    Comparison of different methods to compute $\gS_K$, namely the LogSumExp estimate we use, as well as the Jensen estimate and ratio estimate. As the ratio estimator must work with non-log scale energies $\exp(-\E(x_t))$ it frequently results in NaN estimates which we have omitted from the plot, along with the standard deviations of the ratio estimates (which are also NaN).
    }
    \vspace{-15pt}
    \label{fig:regression_target_analysis}
\end{figure}

We investigate the behaviour of each estimate in Figure \ref{fig:regression_target_analysis}, examining the MSE between the score estimated by each technique and the true score on the GMM task as a function of the number of MC samples. Unfortunately, we observe that despite the Jensen estimate's attractiveness the score estimates it yields have a large MSE which does not readily reduce as the number of MC samples increases indicating a significant bias. On the other hand, the unbiased estimate suffers from extremely large variance due to it having to work directly with the exponentiated energies $\exp(-\E(x_t))$. This results in every estimate before 500 MC samples resulting in exclusively NaN or infinite estimates, rendering the estimate unusable in practice. The only estimate which is well-behaved and approaches the true score is the LogSumExp estimate, whose use we advocate throughout this work. 

Finally, we note that ours is not the only biased objective which achieves good performance in practice: \eg, the current SOTA FAB's objective is also biased due to its use of finite samples for importance sampling as well as its sampling without replacement from the replay buffer.

 \xhdr{Bias vs. Energy for different $K$ and Total Variation Distance}
 
\begin{figure}[h]
    \captionsetup[subfigure]{aboveskip=-1pt,belowskip=-1pt}
    \begin{subfigure}{0.49\linewidth}
        \centering
        \includegraphics[width=\linewidth]{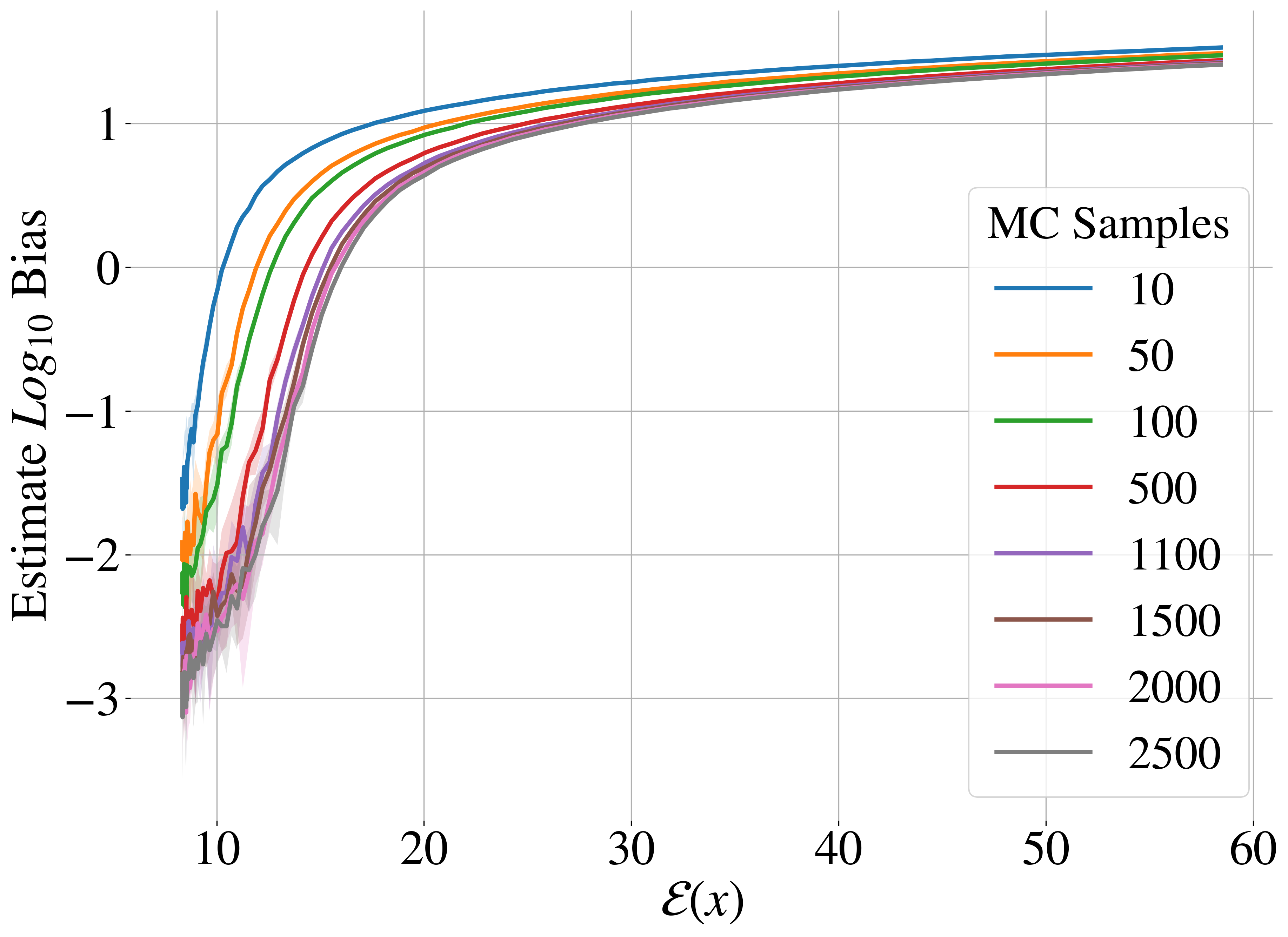}
    \end{subfigure}
    \begin{subfigure}{.49\linewidth}
    \centering
        \includegraphics[width=\linewidth]{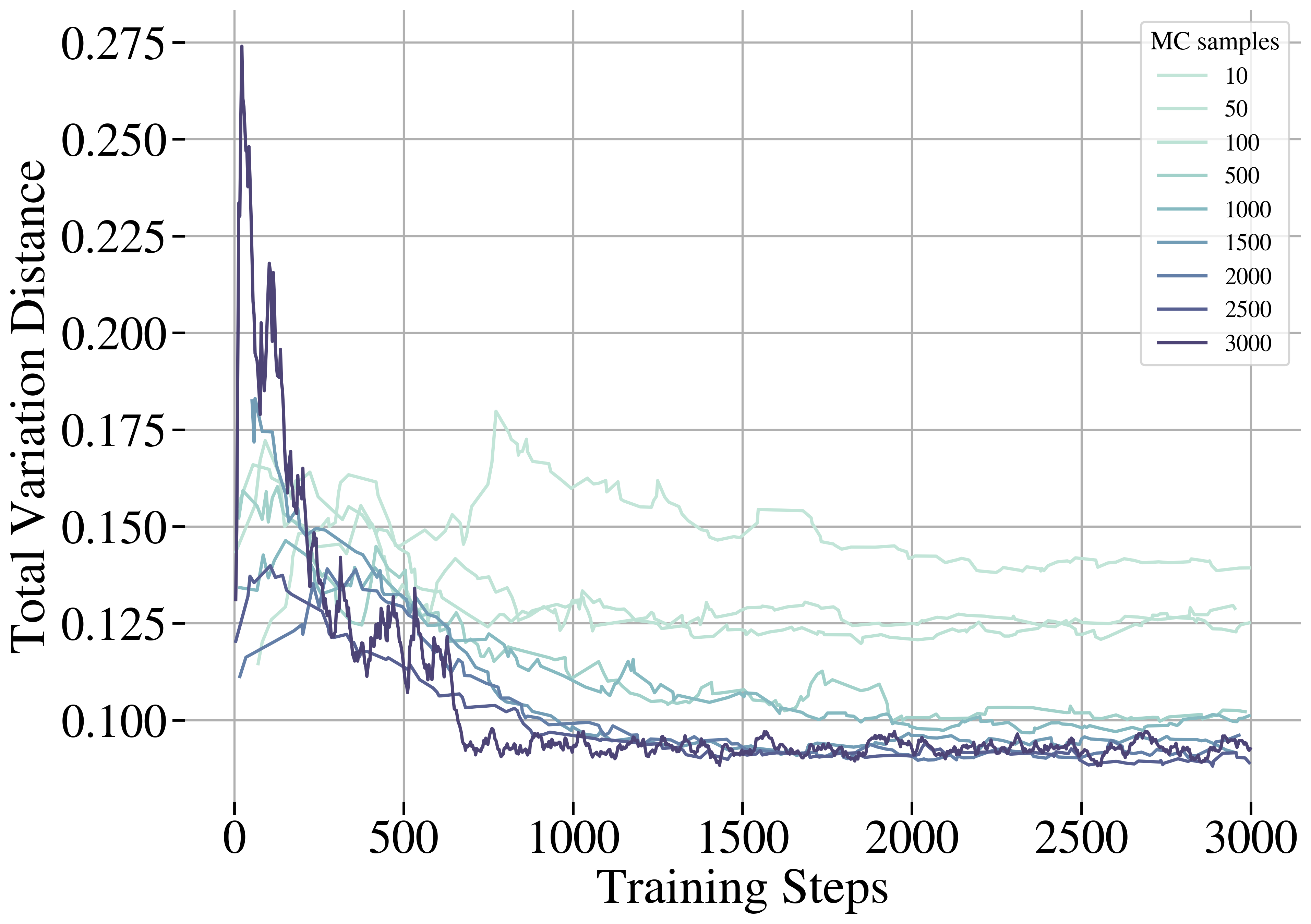}
    \end{subfigure}
    \caption{ \small
       \textbf{Left:} Plot of log bias vs.\ energy for different $K$. The MSE and bias are calculated for GMM with a linear noise schedule. The standard deviations for the log-transformed values are over $10$ seeds with the variance estimated over $256$ samples. For the plot on the right, the values are averaged over $x_0 \sim p_0$ 
    \textbf{Right:} Ablation of Total Variation distance as a function training steps for various numbers of MC samples on DW-4.
    }
    \label{fig:appendix_mc_ablation}
\end{figure}

In \autoref{fig:appendix_mc_ablation} (left) we study the log-bias as a function of $\gE$, by taking a point from a GMM mode and linearly following a direction away from the data itself which corresponds to $\exp(-\gE(x)) \approx 0$. As observed, we notice we need a few MC samples in high-data regions. Additionally, the log-bias increases further from data for a fixed $K$ and drops when we increase $K$. This result empirically confirms \autoref{prop:bias} where we need more MC samples to have a proper estimation of the score.

In \autoref{fig:appendix_mc_ablation} (right) we visualize the progression of the TV metric during training across MC samples on DW-4. In line with expectations, increasing the number of MC samples $K$ leads to a lower biased estimate and a better performance of the model.

\begin{figure}[h]
    \centering
    \includegraphics[width=\linewidth]{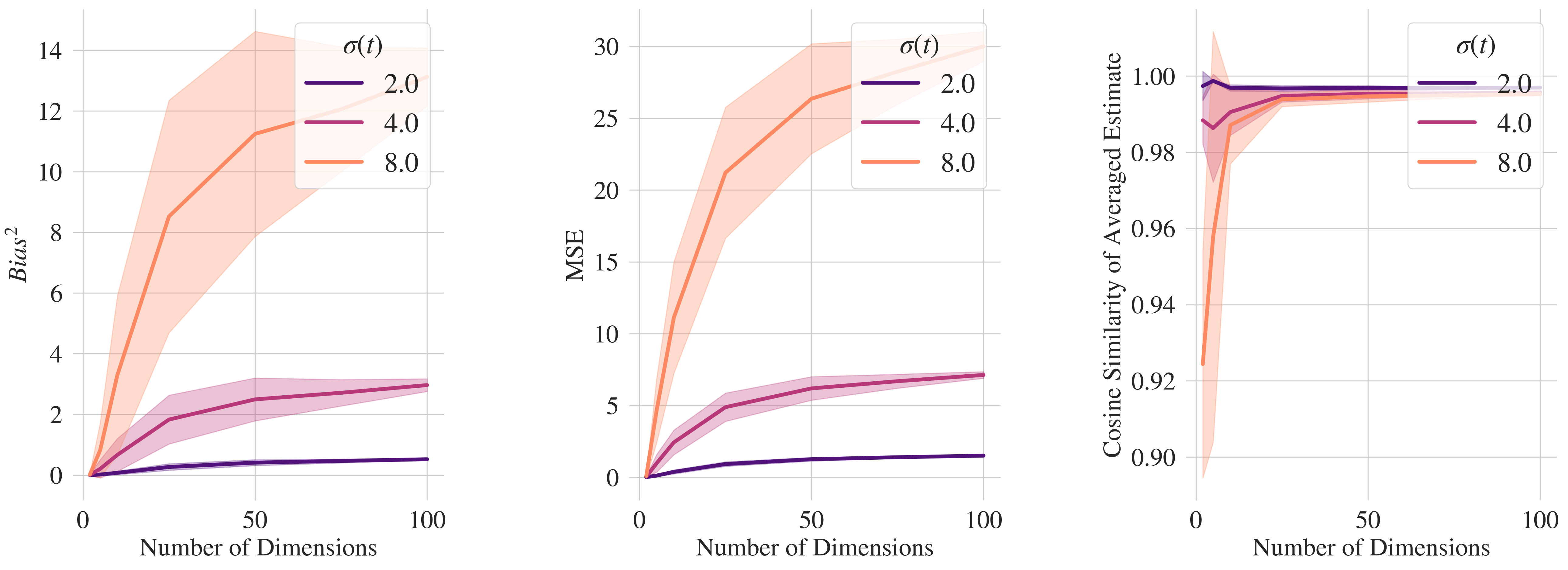}
    \caption{
        Plot of the \textbf{Left:} bias-squared \textbf{Middle:} mean squared error (MSE) and \textbf{Right:} cosine similarity of the mean estimated score to the ground truth score as a function of the number of dimensions of the 40-GMM. These are plotted for $x_t$ sampled at three noise levels $\sigma(t) \in \{2, 4, 8\}$. The standard deviations for all the values are over $10$ seeds with the variance estimated over $128$ samples.}
    \label{fig:gmm_mse_dimensions}
\end{figure}
\xhdr{Estimator quality as a function of dimension}
We examine the quality of the estimator as a function of dimensions, with a closed-form score for the $40$-GMM task. Specifically, we analyze the bias and the MSE (capturing variance) of the estimator as a function of increasing number of dimensions, across various noise levels, $\sigma(t)$ in \autoref{fig:gmm_mse_dimensions} (left, middle). As expected, we find that the bias and MSE increase as a function of dimensions. In line with our expectations and previous findings shown in \autoref{fig:ablation_fig} (right), we note that the bias and MSE are both higher at higher noise levels. In \autoref{fig:gmm_mse_dimensions} (right), we measure the cosine similarity of the mean estimated score to the true score. As this value is very close to $1$ for the dimensions we analyze, we conclude that, interestingly, the estimator error is almost exclusively in the magnitude and not in the direction. This interesting finding is also aligned with our experiments, where we observed that we were able to get good samples, even in high-dimensional tasks, simply by clipping the estimated score. Overall, we hypothesize that overestimating the magnitude of the score early in the diffusion process (i.e., at large values of $\sigma(t)$) does not pose a significant problem as long as the estimated scores still point towards the data density and that later in the diffusion process (i.e., at smaller values of $\sigma(t)$) we are able to get accurate estimates. The reason is that although using the overestimated scores we might move towards the data density too quickly or even overshoot it, by taking small enough steps we will eventually enter a region where our estimates are accurate and we can converge towards the data density.  

\begin{figure}[h]
    \centering
    \begin{subfigure}{0.49\linewidth}
        \centering
        \includegraphics[width=\linewidth]{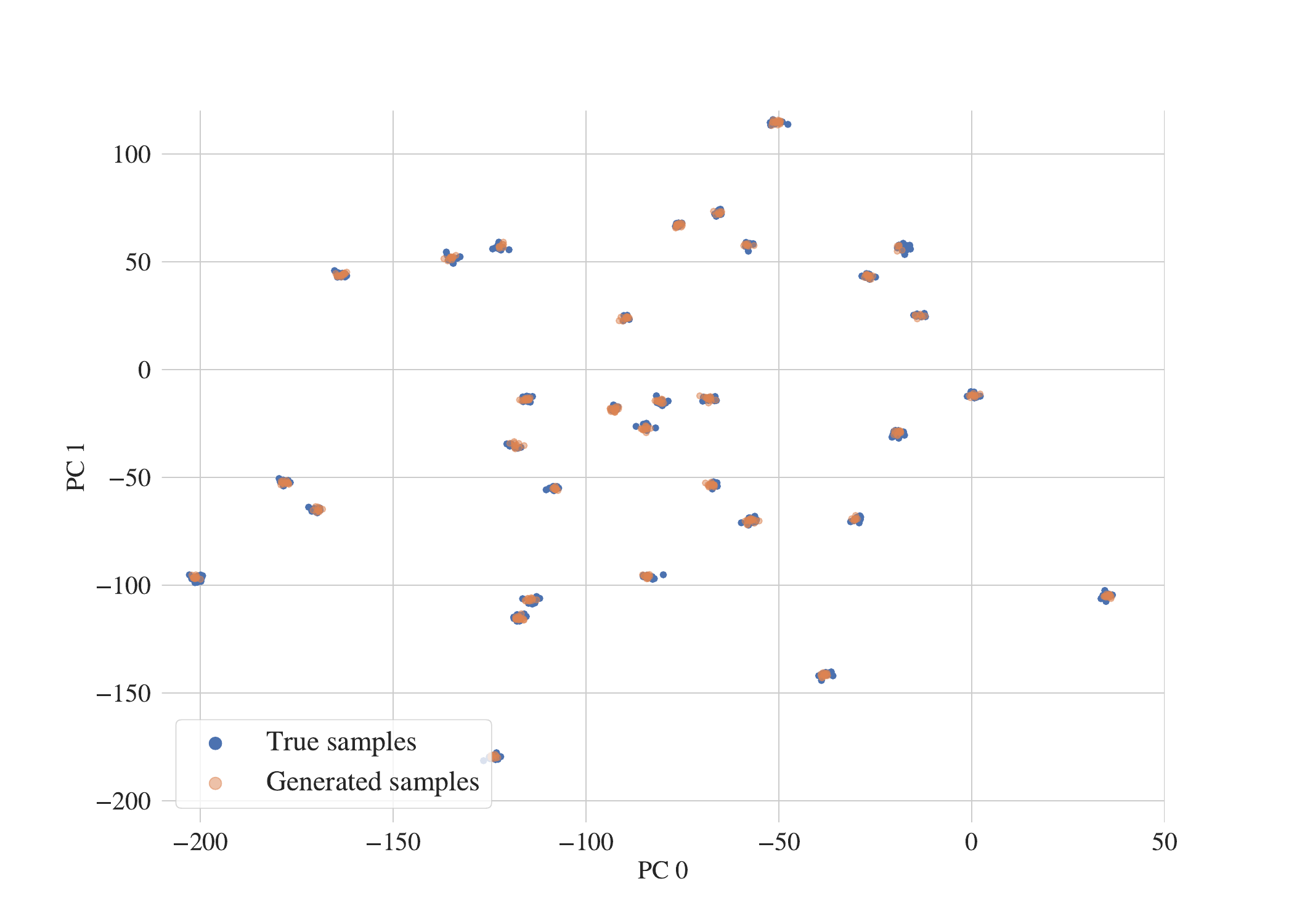}
    \end{subfigure}
    \begin{subfigure}{.49\linewidth}
    \centering
        \includegraphics[width=\linewidth]{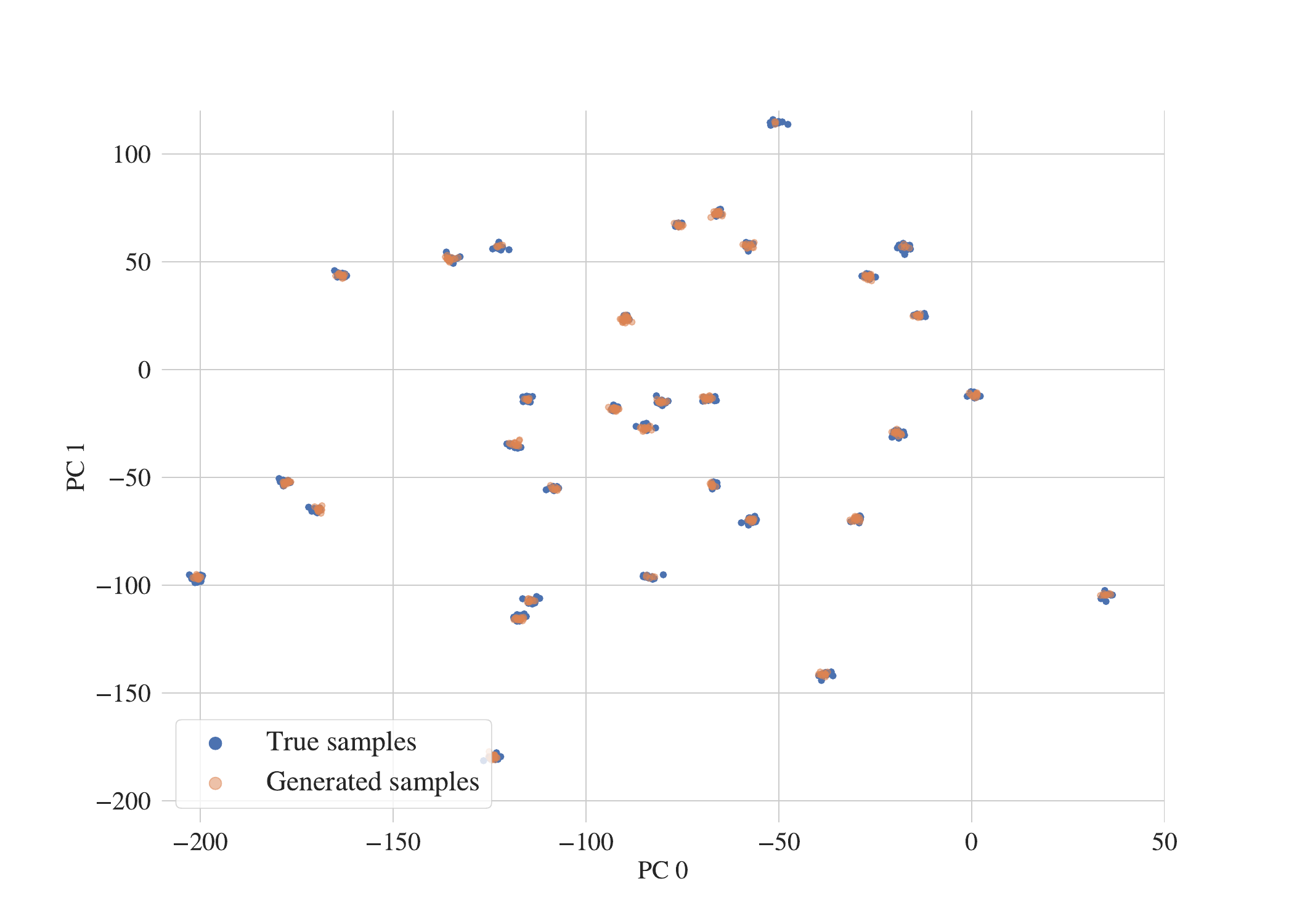}
    \end{subfigure}
    \caption{Generated and true samples for the 40-GMM task in 10,000 dimensions plotted over the first two principal components. The samples are generated using the score estimator with norm clipped to \textbf{Left:} $10^4$ and \textbf{Right:} $10^8$.}
    \label{fig:gmm_samples_10k_pca}
\end{figure}

Finally, to demonstrate that the errors in the magnitude of the estimated score are manageable even in a very high-dimensional setting, we scale up the $40$-GMM task from $2$ to $10000$ dimensions. As we want to examine the quality of the estimator itself, we use the estimator directly, instead of a network, to generate samples. Although clipping is necessary to achieve good samples, the estimator is able to achieve good performance over a wide range of clipping values, from $10^4$ to $10^8$. In \autoref{fig:gmm_samples_10k_pca}, we demonstrate this by plotting the true and predicted distributions on the first two principal components using the estimator, with norm clipped to $10^4$ and $10^8$ respectively.

\end{document}